\pdfoutput=1
\documentclass[10pt,twocolumn,letterpaper]{article}

\usepackage{cvpr}
\usepackage{times}
\usepackage{epsfig}
\usepackage{graphicx}
\usepackage{amsmath}
\usepackage{amssymb}
\usepackage[breaklinks=true,bookmarks=false]{hyperref}
\usepackage{capt-of}
\usepackage{color}
\usepackage{subcaption}
\usepackage{algpseudocode}
\usepackage{algorithm}
\usepackage{multirow}

\algtext*{EndIf}

\DeclareMathOperator*{\argmax}{\arg\!\max}

\cvprfinalcopy % *** Uncomment this line for the final submission

\begin{document}

%%%%%%%%% TITLE
\title{Modeling and Propagating CNNs in a Tree Structure for Visual Tracking}

\author{Hyeonseob Nam\thanks{Both authors have equal contribution on this paper.}~~~~~~~~Mooyeol Baek\footnotemark[1]~~~~~~~~Bohyung Han\\
Department of Computer Science and Engineering, POSTECH, Korea\\
{\tt\small \{namhs09, mooyeol, bhhan\}@postech.ac.kr}
}

\maketitle
%%%%%%%%% ABSTRACT
\begin{abstract}
We present an online visual tracking algorithm by managing multiple target appearance models in a tree structure.
The proposed algorithm employs Convolutional Neural Networks (CNNs) to represent target appearances, where multiple CNNs collaborate to estimate target states and determine the desirable paths for online model updates in the tree.
By maintaining multiple CNNs in diverse branches of tree structure, it is convenient to deal with multi-modality in target appearances and preserve model reliability through smooth updates along tree paths.
Since multiple CNNs share all parameters in convolutional layers, it takes advantage of multiple models with little extra cost by saving memory space and avoiding redundant network evaluations.
The final target state is estimated by sampling target candidates around the state in the previous frame and identifying the best sample in terms of a weighted average score from a set of active CNNs.
Our algorithm illustrates outstanding performance compared to the state-of-the-art techniques in challenging datasets such as online tracking benchmark and visual object tracking challenge.
\end{abstract}

%%%%%%%%% BODY TEXT
%---------------------------------------------------------------------------------------------------------------------------------------------------------
% INTRODUCTION
%---------------------------------------------------------------------------------------------------------------------------------------------------------
\section{Introduction}
\label{sec:introduction}
Visual tracking is a fundamental computer vision task with a wide range of applications.
Among many subproblems in visual tracking, target appearance modeling is one of the most critical components and there are various approaches to maintain adaptive and robust models.
This problem is typically formulated as an online learning framework, where a generative or discriminative model is incrementally updated during tracking to adapt the variations in target appearances.
Most existing tracking algorithms update the target model assuming that target appearances change smoothly over time.
However, this strategy may not be appropriate for handling more challenging situations such as occlusion, illumination variation, abrupt motion and deformation, which may break temporal smoothness assumption.
Some algorithms employ multiple models~\cite{zhang2014meem,mei2009robust,zhang2012robust}, multi-modal representations~\cite{han2005online} or nonlinear classifiers~\cite{grabner2006real,hare2011struck} to address these issues.
However, the constructed models are still not strong enough and online model updates are limited to sequential learning in a temporal order, which may not be able to make the models sufficiently discriminative and diverse.

CNNs have recently gained significant attention for various visual recognition problems such as image classification~\cite{krizhevsky2012imagenet}, object detection and localization~\cite{girshick2014rich}, image segmentation~\cite{long2014fully,noh2015learning} due to their  strong representation power.
However, online learning with CNNs is not straightforward because neural networks tend to forget previously learned information quickly when they learn new information~\cite{mccloskey1989catastrophic}.
This property often incurs drift problem especially when background information contaminates target appearance models, targets are completely occluded by other objects, or tracking fails temporarily.
This problem may be alleviated by maintaining multiple versions of target appearance models constructed at different time steps and updating a subset of models selectively to keep a history of target appearances.
This idea has been investigated in \cite{li122014deeptrack}, where a pool of CNNs are used to model target appearances, but it does not consider the reliability of each CNN to estimate target states and update models.

We propose an online visual tracking algorithm, which estimates target state using the likelihoods obtained from multiple CNNs.
The CNNs are maintained in a tree structure and updated online along the path in the tree.
Since each path keeps track of a separate history about target appearance changes, the proposed algorithm is effective to handle multi-modal target appearances and other exceptions such as short-term occlusions and tracking failures.
In addition, since the new model corresponding to the current frame is constructed by fine-tuning the CNN that produces the highest likelihood for target state estimation, more consistent and reliable models are to be generated through online learning only with few training examples.

The main contributions of our paper are summarized below:
\begin{itemize}
\item[$\bullet$] We propose a visual tracking algorithm to manage target appearance models based on CNNs in a tree structure, where the models are updated online along the path in the tree.
This strategy enables us to learn more persistent models through smooth updates.
\item[$\bullet$] Our tracking algorithm employs multiple models to capture diverse target appearances and performs more robust tracking even with challenges such as appearance changes, occlusions, and temporary tracking failures.
\item[$\bullet$] The proposed algorithm presents outstanding accuracy in the standard tracking benchmarks~\cite{otb1,otb2,vot14} and outperforms the state-of-the-art methods by large margins.
\end{itemize}

The rest of this paper is organized as follows.
We first review various techniques in visual tracking in Section~\ref{sec:related}, and then describe the overview of the proposed approach in Section~\ref{sec:algorithm}. 
Section~\ref{sec:proposed} discusses the detailed algorithm and Section~\ref{sec:experiment} illustrates experimental results in multiple challenging datasets.

%---------------------------------------------------------------------------------------------------------------------------------------------------------
% RELATED WORK
%---------------------------------------------------------------------------------------------------------------------------------------------------------
% !TEX root = eccv16_tcnn.tex

\section{Related Work}
\label{sec:related}
There are various kinds of visual tracking algorithms, and it is difficult to review all the prior works. 
In this section, we focus on several discriminative tracking algorithms based on tracking-by-detection first, and then discuss more specific topics such as visual tracking with multiple target appearance models and representation learning based on CNNs for visual tracking.

Tracking-by-detection approaches formulate visual tracking as a discriminative object classification problem in a sequence of video frames.
The techniques in this category typically learn classifiers to differentiate targets from surrounding backgrounds; various algorithms have achieved improved performance by coping with dynamic appearance changes and constructing robust target models.
For example, \cite{grabner2006real} modified a famous object detection algorithm, Adaboost, and presented an online learning method for tracking. 
A multiple instance learning technique has been introduced in \cite{BabenkoTPAMI11} to update classifier online, where a bag of image patches is employed as a training example instead of a single patch to alleviate labeling noises.
By the similar motivation, an approach based on structured SVM has been proposed in \cite{hare2011struck}.
TLD~\cite{kalal2012tracking} proposed a semi-supervised learning technique with structural constraints.
All of these techniques are successful in learning reasonable target representations by adopting online discriminative learning procedures, but still rely on simple shallow features; we believe that tracking performance may be improved further by using deep features.

Although the representations by deep neural networks turn out to be effective in various visual recognition problems, tracking algorithms based on hand-crafted features~\cite{hong2015multi,zhang2014meem} often outperform CNN-based approaches.
This is partly because CNNs are difficult to train using noisy labeled data online while they are easy to overfit to a small number of training examples; it is not straightforward to apply CNNs to visual tracking problems involving online learning.
For example, the performance of \cite{li122014deeptrack}, which is based on a shallow custom neural network, is not as successful as recent tracking algorithms based on shallow feature learning.
However, CNN-based tracking algorithms started to present competitive accuracy in the online tracking benchmark~\cite{otb1} by transferring the CNNs pretrained on ImageNet~\cite{ImageNet}.
For example, simple approaches based on fully convolutional networks or hierarchical representations illustrate substantially improved results~\cite{ma2015hierarchical,wang2015visual}.
In addition, the combination of pretrained CNN and online SVM achieves competitive results~\cite{hong2015online}.
However, these deep learning based methods are still not very impressive compared to the tracking techniques based on hand-crafted features~\cite{hong2015multi}.

Multiple models are often employed in generative tracking algorithms to handle target appearance variations and recover from tracking failures.
Trackers based on sparse representation~\cite{mei2009robust,zhang2012robust} maintain multiple target templates to compute the likelihood of each sample by minimizing its reconstruction error while \cite{kwon2010visual} integrates multiple observation models via an MCMC framework.
Nam~\etal~\cite{nam2014online} integrates patch-matching results from multiple frames and estimates the posterior of target state.
On the other hand, ensemble classifiers have sometimes been applied to visual tracking problem.
Tang~\etal~\cite{tang2007co} proposed a co-tracking framework based on two support vector machines.
An ensemble of weak classifiers is employed to estimate target states in \cite{Avidan2007ensemble,Bai2013randomized}.
Zhang~\etal~\cite{zhang2014meem} presented a framework based on multiple snapshots of SVM-based trackers to recover from tracking failures.

%---------------------------------------------------------------------------------------------------------------------------------------------------------
% Algorithm Overview
%---------------------------------------------------------------------------------------------------------------------------------------------------------
\section{Algorithm Overview}
\label{sec:algorithm}
\begin{figure}[t]
\begin{center}
\begin{subfigure}[b]{0.49\linewidth}
\centering
\includegraphics[width=0.7\linewidth]{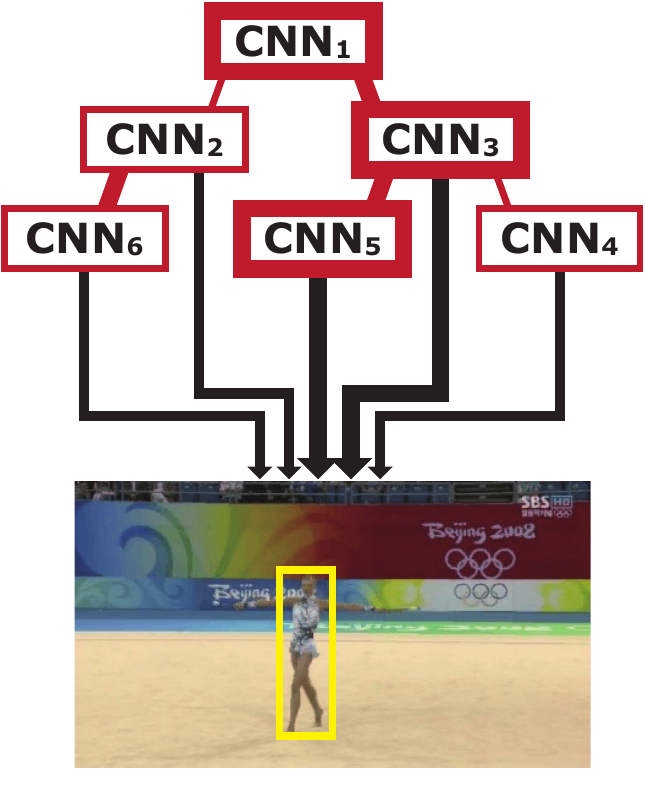}
\caption{State estimation}
\label{fig:overview1}
\end{subfigure}
\begin{subfigure}[b]{0.49\linewidth}
\centering
\includegraphics[width=0.7\linewidth]{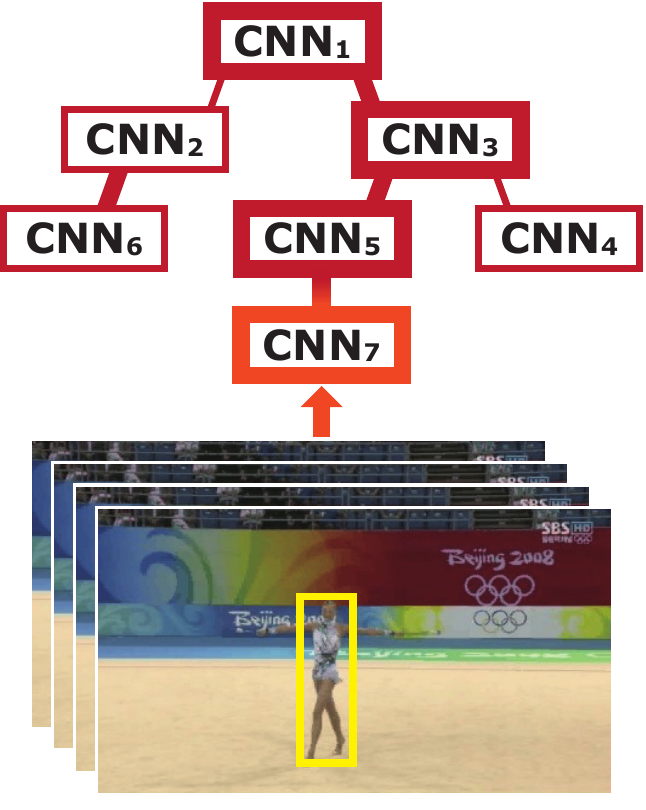}
\caption{Model update}
\label{fig:overview2}
\end{subfigure}
\end{center}
\vspace{-2mm}
\caption{Illustration of target state estimation and model update procedures using multiple CNNs maintained in a tree structure. The width of a black arrow indicates the weight of a CNN for target state estimation while the width of a red edge denotes the affinity between two CNNs. The width of box outline means the reliability of the CNN associated with the box.}
\label{fig:overview}
\end{figure}

Our algorithm maintains multiple target appearance models based on CNNs in a tree structure to preserve model consistency and handle appearance multi-modality effectively.
The proposed approach consists of two main components as in ordinary tracking algorithms---state estimation and model update---whose procedures are illustrated in Figure~\ref{fig:overview}.
Note that both components require interaction between multiple CNNs.

When a new frame is given, we draw candidate samples around the target state estimated in the previous frame, and compute the likelihood of each sample based on the weighted average of the scores from multiple CNNs.
The weight of each CNN is determined by the reliability of the path along which the CNN has been updated in the tree structure.
The target state in the current frame is estimated by finding the candidate with the maximum likelihood.
After tracking a predefined number of frames, a new CNN is derived from an existing one, which has the highest weight among the contributing CNNs to target state estimation.
This strategy is helpful to ensure smooth model updates and maintain reliable models in practice.

Our approach has something in common with \cite{li122014deeptrack}, which employs a candidate pool of multiple CNNs. 
It selects $k$ nearest CNNs based on prototype matching distances for tracking.
Our algorithm is differentiated from this approach since it is more interested in how to keep multi-modality of multiple CNNs and maximize their reliability by introducing a novel model maintenance technique using a tree structure.
Visual tracking based on a tree-structured graphical model has been investigated in \cite{hong2014visual}, but this work is focused on identifying the optimal density propagation path for offline tracking.
The idea in \cite{nam2014online} is also related, but it mainly discusses posterior propagation on directed acyclic graphs for visual tracking.

%---------------------------------------------------------------------------------------------------------------------------------------------------------
% Main Algorithm
%---------------------------------------------------------------------------------------------------------------------------------------------------------
\section{Proposed Algorithm}
\label{sec:proposed}
This section describes the architecture of our CNN employed to learn discriminative target appearance models.
Then, we discuss the detailed procedure of our tracking algorithm, which includes the method to maintain multiple CNNs in a tree structure for robust appearance modeling.

%-------------------------------------------------------------------------
\subsection{CNN Architecture}
\begin{figure*}[t]
\begin{center}
\includegraphics[width=1\linewidth]{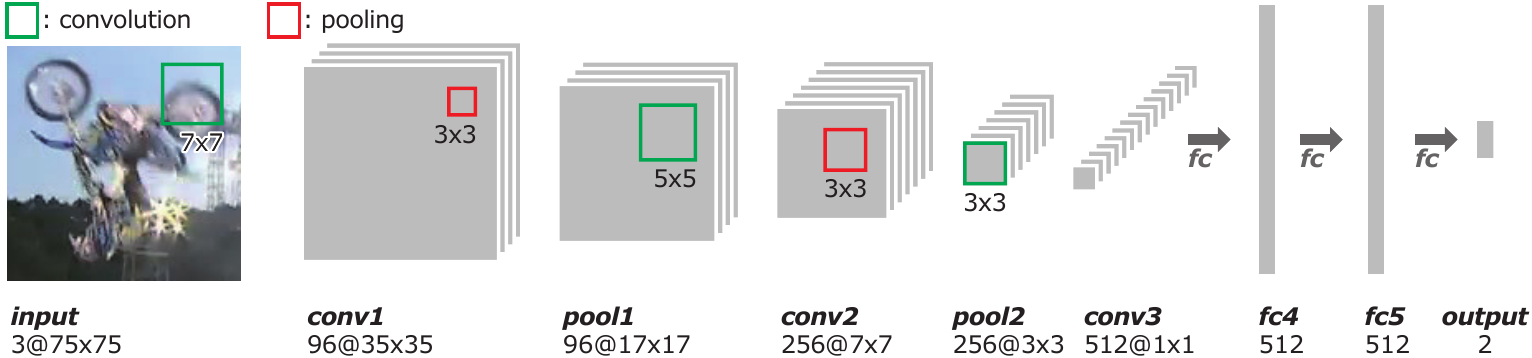}
\end{center}
   \caption{The architecture of our network. We transfer VGG-M network pretrained on ImageNet for convolutional layers while all the fully connected layers are initialized randomly.  The number of channels and the size of each feature map are shown with the name of each layer.}
\label{fig:arch}
\end{figure*}

Our network consists of three convolutional layers and three fully connected layers.
The convolution filters are identical to the ones in VGG-M network~\cite{chatfield2014return} pretrained on ImageNet~\cite{ImageNet}.
The last fully connected layer has 2 units for binary classification while the preceding two fully connected layers are composed of 512 units.
All weights in these three layers are initialized randomly.
The input to our network is a $75\times75$ RGB image and its size is equivalent to the receptive field size of the only single unit (per channel) in the last convolutional layer.
Note that, although we borrow the convolution filters from VGG-M network, the size of our network is smaller than the original VGG-M network.
The output of an input image $\mathbf{x}$ is a normalized vector $[\phi(\mathbf{x}), 1-\phi(\mathbf{x})]^\text{T}$, whose elements represent scores for target and background, respectively.
The overall architecture of our CNN is illustrated in Figure~\ref{fig:arch}.

%-------------------------------------------------------------------------
\subsection{Tree Construction}
We maintain a tree structure to manage hierarchical multiple target appearance models based on CNNs.
In the tree structure $\mathcal{T}=\{\mathcal{V},\mathcal{E}\}$, a vertex $v\in\mathcal{V}$ corresponds to a CNN and a directed edge $(u,v)\in\mathcal{E}$ defines the relationship between CNNs.
The score of an edge $(u,v)$ is the affinity between two end vertices, which is given by
\begin{equation}
\label{eq:edge}
s(u,v)=\frac{1}{|\mathcal{F}_v|}\sum_{t\in\mathcal{F}_v} \phi_u(x_t^*),
\end{equation}
where $\mathcal{F}_v$ is a set of consecutive frames that is used to train the CNN associated with $v$, $x_t^*$ is the estimated target state at frame $t$, and $\phi_u(\cdot)$ is the predicted positive score with respect to the CNN in $u$.
Note that the edge scores play crucial roles in estimating target states and providing reliable paths for model update.
The details about this issue are discussed next.

%-------------------------------------------------------------------------
\subsection{Target State Estimation using Multiple CNNs}
The proposed algorithm estimates the target state in a new frame by aggregating the scores from multiple CNNs in the tree structure.
Let $\mathbf{x}_t^1,\dots,\mathbf{x}_t^N$ be $N$ target candidates in the current frame $t$, which are sampled around the previous target state, and $\mathcal{V}_{+} \subseteq \mathcal{V}$ be the active set of CNNs, which contribute to the state estimation.
The target score $H(\mathbf{x}_t^i)$ for sample $\mathbf{x}_t^i$ is computed by a weighted average of CNN scores, which is given by `
\begin{equation}
\label{eq:aggregation}
H(\mathbf{x}_t^i) = \sum_{v\in\mathcal{V}_{+}} w_{v\rightarrow t} \phi_v(\mathbf{x}_t^i),
\end{equation}
where $w_{v\rightarrow t}$ denotes the weight of the CNN corresponding to vertex $v$ for tracking target in frame $t$.
The optimal target state $\mathbf{x}_t^*$ is given by the candidate sample with the maximum target score as
\begin{equation}
\label{eq:state}
\mathbf{x}_t^* = \arg\!\max_{\mathbf{x}_t^i} H(\mathbf{x}_t^i).
\end{equation}

The remaining issue is how to determine the weight $w_{v\rightarrow t}$ in Eq.~\eqref{eq:aggregation}.
The weight is identified by the following two factors: affinity to the current frame and reliability of CNN.
The affinity $\alpha_{v\rightarrow t}$ indicates how confidently a CNN in vertex $v$ affects the tracking result in frame $t$, which is determined by the maximum positive score over all candidates as follows:  
\begin{equation}
\label{eq:affinity}
\alpha_{v\rightarrow t}=\max_{\mathbf{x}_t^i} \phi_v(\mathbf{x}_t^i).
\end{equation}
However, the estimation of the weight $w_{v\rightarrow t}$ based only on this measure may be problematic because it ignores how reliable each CNN is.
For example, if a CNN is fine-tuned on the frames with complete failures or nontrivial errors,  the CNN is likely to produce high scores for background objects and should be penalized despite the existence of samples with high affinities.\footnote{We still need to maintain such CNNs since we sometimes need to follow background objects to estimate target state in case of severe occlusion.}

To address this issue, we also employ the reliability of each CNN for target state estimation using the path in the tree structure along which the CNN has been updated.
Note that we may have an unreliable CNN in the path, which may has been updated using the frames with tracking errors.
We estimate the reliability of CNN associated with vertex $v$, which is denoted by $\beta_v$, using the score in the bottleneck edge.
Specifically, the reliability of a CNN is efficiently computed in a recursive manner without exploring the entire path, which is formally given by
\begin{equation}
\label{eq:reliability}
\beta_v = \min\left(s(p_v,v),\beta_{p_v} \right),
\end{equation}
where $p_v$ is the parent of CNN in vertex $v$, which has an outgoing edge $(p_v,v)$ in the tree structure.

Based on these two criteria, the combined weight of the CNN corresponding to vertex $v$, $w_{v\rightarrow t}$, is given by 
\begin{equation}
\label{eq:weight}
w_{v\rightarrow t} = \frac{\min(\alpha_{v\rightarrow t},\beta_v)}{\sum_{v \in \mathcal{V}_{+}} \min(\alpha_{v\rightarrow t},\beta_v)}.
\end{equation}
Note that both $\alpha_{v \rightarrow t}$ and $\beta_v$ are the scores evaluated on CNNs, and taking the minimum value out of two bottleneck similarities determines the weight of a new CNN associated with frame $t$ when it is updated from CNN in vertex $v$.

%-------------------------------------------------------------------------
\subsection{Bounding Box Regression}
The target state estimated in Eq.~\eqref{eq:state} may not correspond to the tight bounding box since the representation by CNN is not appropriate for accurate localization of an object.
Therefore, we employ the bounding box regression technique, which is frequently used in object detection~\cite{girshick2014rich}, to enhance target localization quality.
We learn a simple regression function at the first frame, and adjust target bounding boxes using the model in the subsequent frames.
This idea is simple but effective to improve performance.
Please, refer to \cite{girshick2014rich} for details.

%-------------------------------------------------------------------------
\subsection{Model Update}
We maintain multiple CNNs in a tree structure, where a CNN is stored in each node.
This approach improves multi-modality of appearance models by having CNNs in diverse branches while it preserves model reliability by smoothly updating CNNs along the path in the tree.
Now, the critical issue is how to select the appropriate path for fine-tuning a CNN using new training examples.

We create a node $z$ for the new CNN after finishing tracking $\Delta(=10)$ consecutive frames without model updates, which are elements in $\mathcal{F}_z$.
The parent CNN is the one to maximize the reliability of the new CNN, and the vertex $p_z$ associated with the parent CNN is identified by
\begin{equation}
\label{eq:parent}
p_z = \argmax_{v\in\mathcal{V}_{+}} \min(\tilde{s}(v,z), \beta_v),
\end{equation}
where $\tilde{s}(v,z)$ indicates the score of a tentative edge $(v,z)$, which is computed by Eq.~\eqref{eq:edge}.
The CNN in vertex $z$ is fine-tuned from the CNN in $p_z$ using the training samples collected from two sets of frames, $\mathcal{F}_z$ and $\mathcal{F}_{p_z}$.
The tree structure is expanded by adding a new vertex $z$ and the corresponding edge $(p_z, z)$.
The active CNN set denoted by $\mathcal{V}_{+}$ is also updated to contain $K(=10)$ CNNs, which are the ones added to the tree structure most recently.

%-------------------------------------------------------------------------
\subsection{Implementation Details}

% State estimation
To estimate the target state in each frame, we draw 256 samples in $(x,y,s)$ space from an independent multivariate normal distribution centered at the previous target state.
The standard deviations of the three variables are given by $\sigma_x=\sigma_y=0.3l$ and $\sigma_s=0.5$; $l$ is the average of width and height of the previous target bounding box, and the scale of a sample bounding box is determined by multiplying $1.05^{s}$ to the width and height of the initial target.

% Model update
We only update the fully connected layers while all the convolutional layers are identical to the original pretrained network on ImageNet.
Although we store multiple CNNs in the tree structure, the parameters of all convolutional layers are shared and the outputs from the shared layers can be reused.
Therefore, maintaining and evaluating multiple CNNs require little additional cost compared to handling a single CNN in terms of time and space complexity.

The training data are collected whenever tracking is completed in each frame, where we draw 50 positive and 200 negative examples that have larger than 0.7 IoU and less than 0.5 IoU with the estimated target bounding boxes, respectively.
In practice, we do not store image patches as training examples but save their {\sf conv$_3$} features since the features in the layer do not change and it is not necessary to perform convolution operations more than once.

% Network parameters
Our CNNs are trained by the standard Stochastic Gradient Descent (SGD) method with softmax cross-entropy loss.
They are trained for 10 iterations with learning rate 0.003, while the initial CNN is trained for 50 iterations with learning rate 0.001.
The weight decay and momentum parameters are given by 0.0005 and 0.9, respectively.
All parameters are fixed throughout our experiment.

%---------------------------------------------------------------------------------------------------------------------------------------------------------
% EXPERIMENTS
%---------------------------------------------------------------------------------------------------------------------------------------------------------
\section{Experiment}
\label{sec:experiment}

Our algorithm is implemented in {\sf MATLAB} using {\sf MatConvNet} library~\cite{vedaldi15matconvnet}.
The average speed is approximately 1.5 fps without optimization using an {\sf Intel Core i7-5820K} CPU with 3.30GHz and a single {\sf NVIDIA GeForce GTX TITAN X} GPU.
The proposed algorithm is tested on two standard datasets for tracking performance evaluation; one is online tracking benchmark (OTB)~\cite{otb1,otb2} and the other is visual object tracking 2015 benchmark (VOT2015)~\cite{vot15}.

\subsection{Evaluation on OTB dataset}
We first describe the evaluation results of our algorithm on online tracking benchmark~\cite{otb1,otb2}.
OTB50~\cite{otb1} contains 50 fully-annotated video sequences with various targets and backgrounds, and its extension, denoted by OTB100~\cite{otb2}, has additional 50 video sequences.
In addition to target bounding box ground-truths, each sequence has annotations for its own challenging attributes such as occlusion, background clutter, illumination change, motion blur, etc.
This information is useful to analyze the characteristics of individual trackers.

\subsubsection{Evaluation Methodology}
We follow the evaluation protocol provided by the benchmark~\cite{otb1,otb2}, where the performance of trackers is evaluated based on two criteria: bounding box overlap ratio and center location error. 
The success and precision plots are generated by computing the rates of successfully tracked frames at many different thresholds in the two criteria.
The ranks of trackers are determined by the accuracy at 20 pixel threshold in the precision plot and the Area Under Curve (AUC) score in the success plot. 

\subsubsection{Results}
\begin{figure}[t]
\begin{center}
\begin{subfigure}{\linewidth}
\centering
\includegraphics[width=0.495\linewidth]{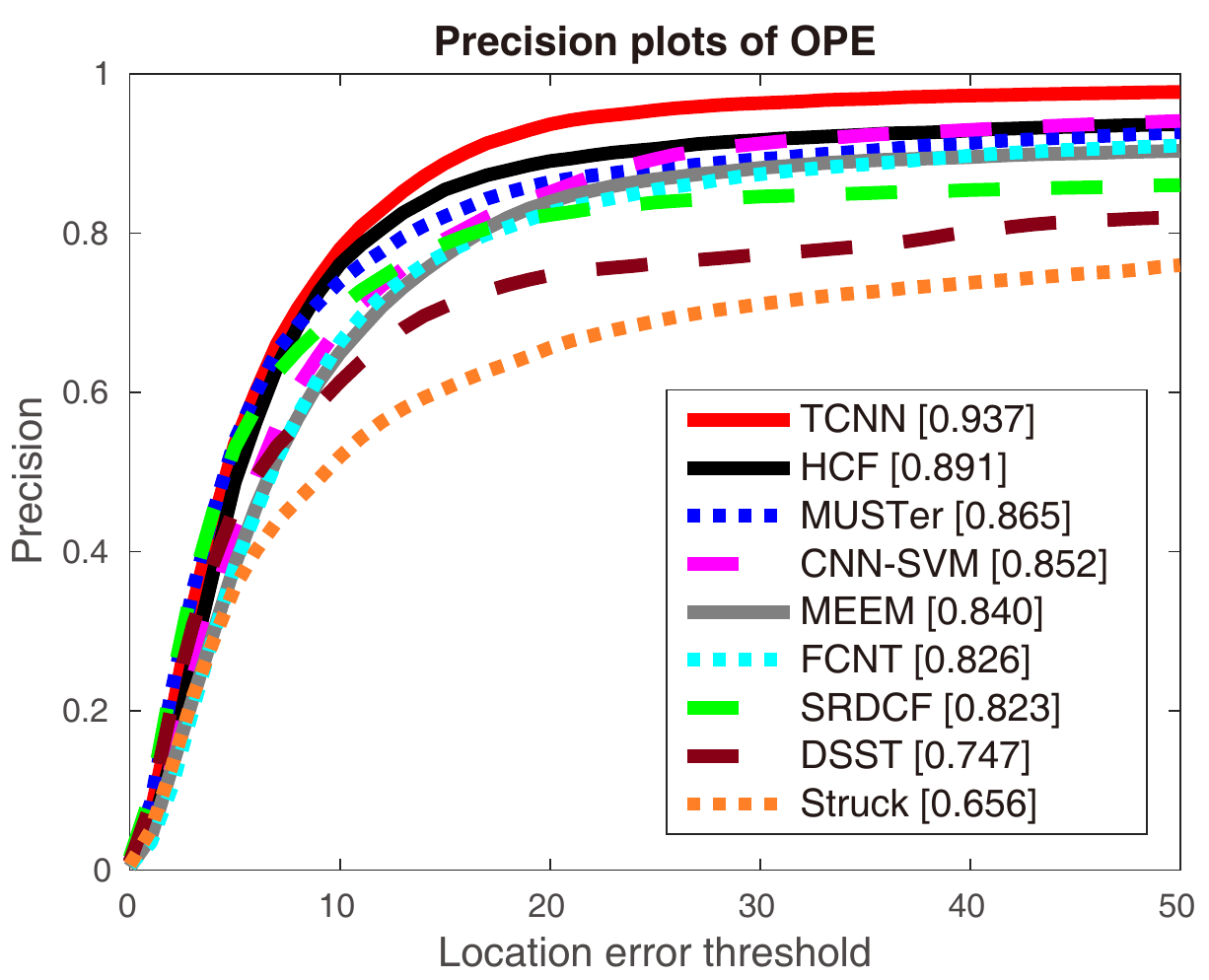}
\includegraphics[width=0.495\linewidth]{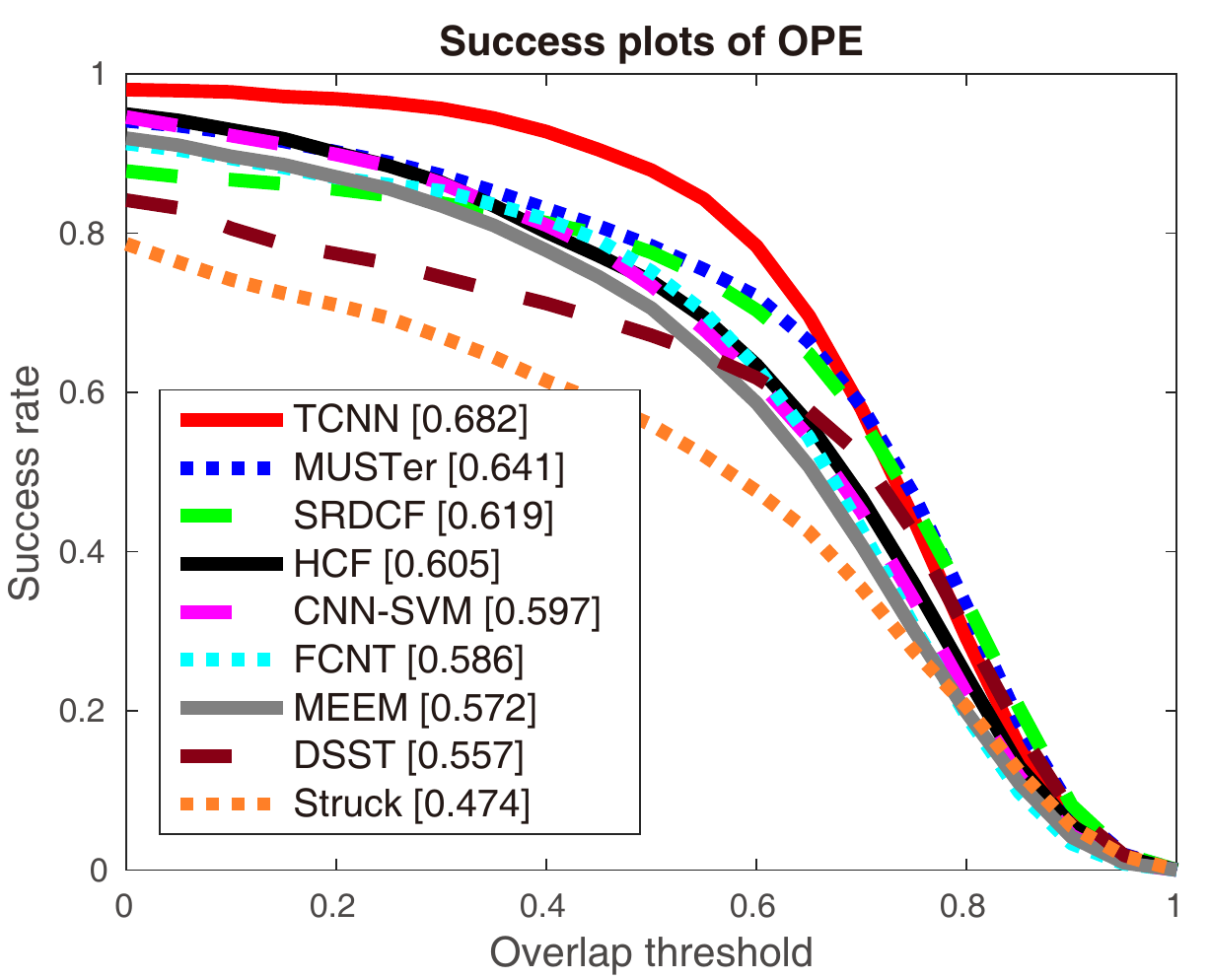}
\caption{Results for OTB50 sequences in \cite{otb1}}
\label{fig:otb50}
\end{subfigure}
\begin{subfigure}{\linewidth}
\centering
\includegraphics[width=0.495\linewidth]{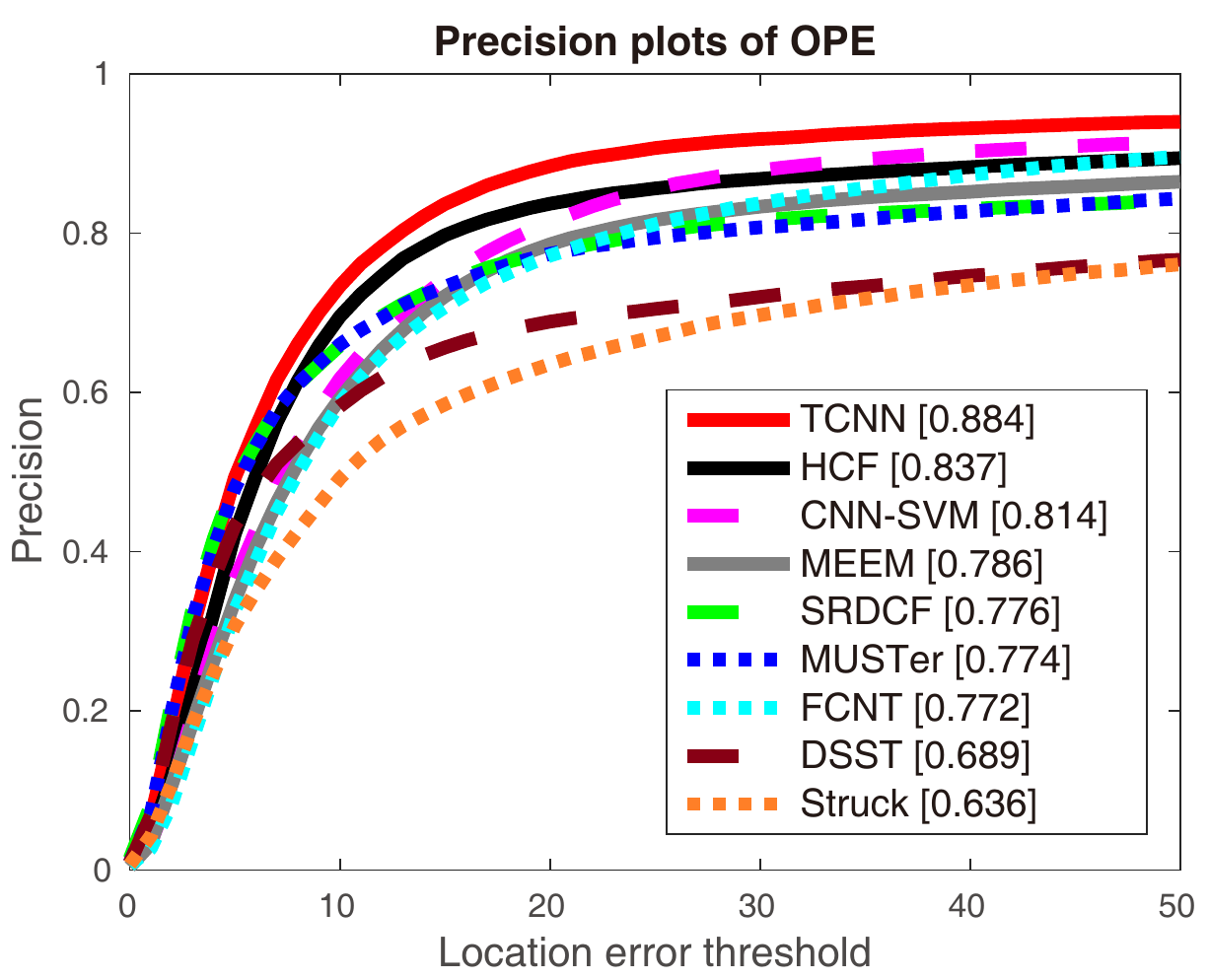}
\includegraphics[width=0.495\linewidth]{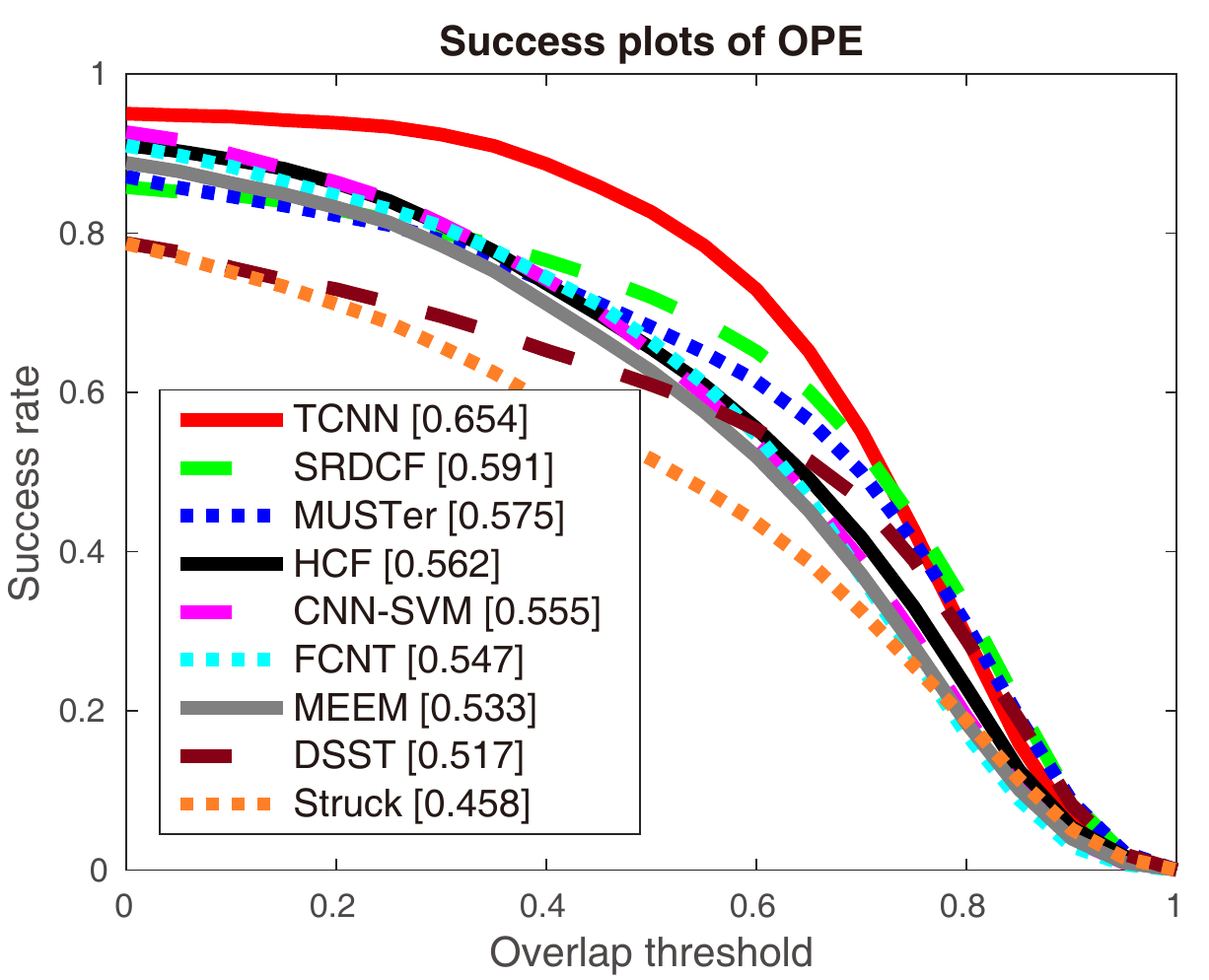}
\caption{Results for OTB100 sequences in \cite{otb2}}
\label{fig:otb100}
\end{subfigure}
\end{center}
\vspace{-5mm}
\caption{Quantitative results on the benchmark datasets. The values in the legend are the precision at a threshold of 20 pixels and the AUC score, for precision and success plots, respectively.}
\label{fig:otb_result}
\end{figure}

\begin{figure*}[t]
\begin{center}
\includegraphics[width=0.24\linewidth]{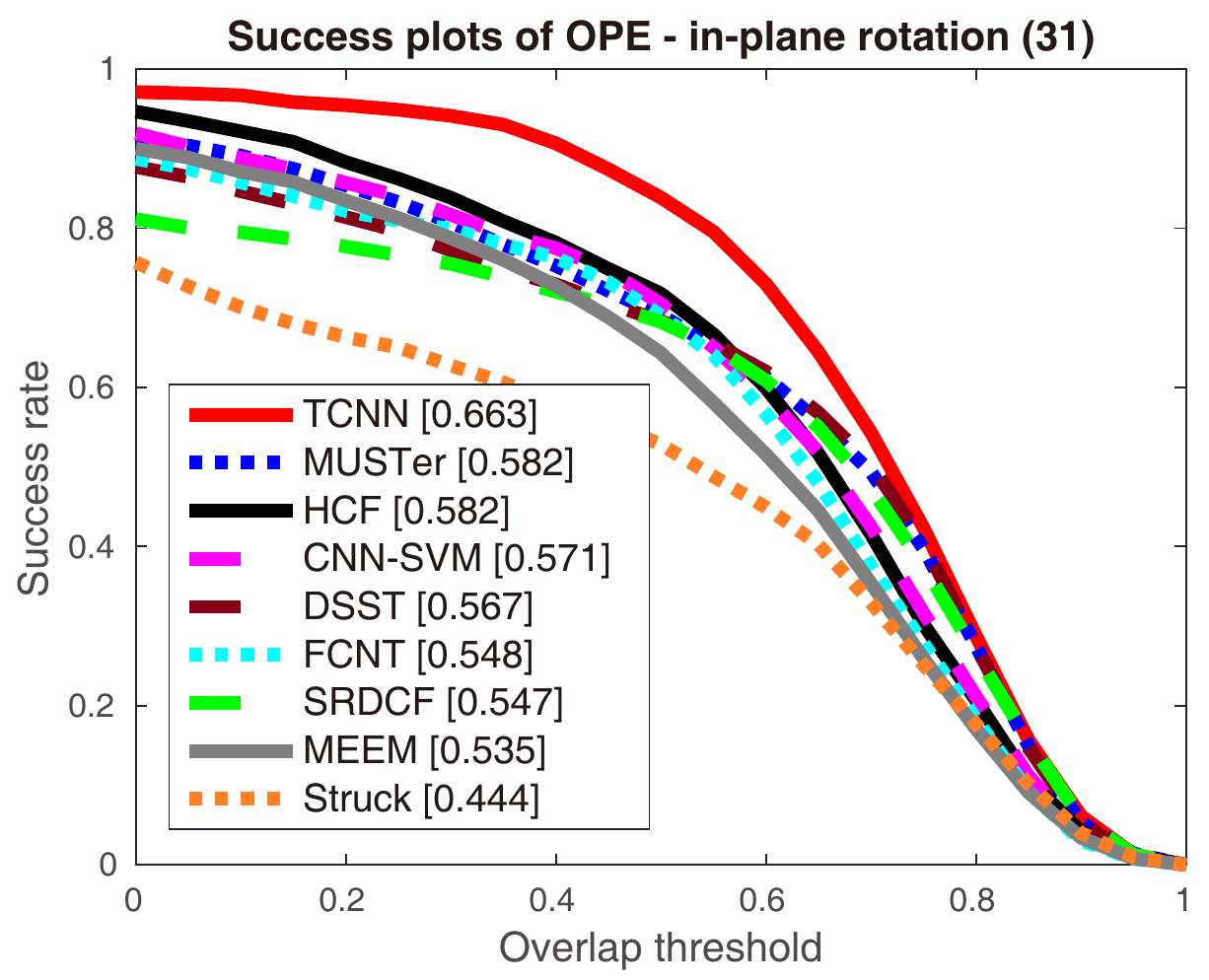}%
\includegraphics[width=0.24\linewidth]{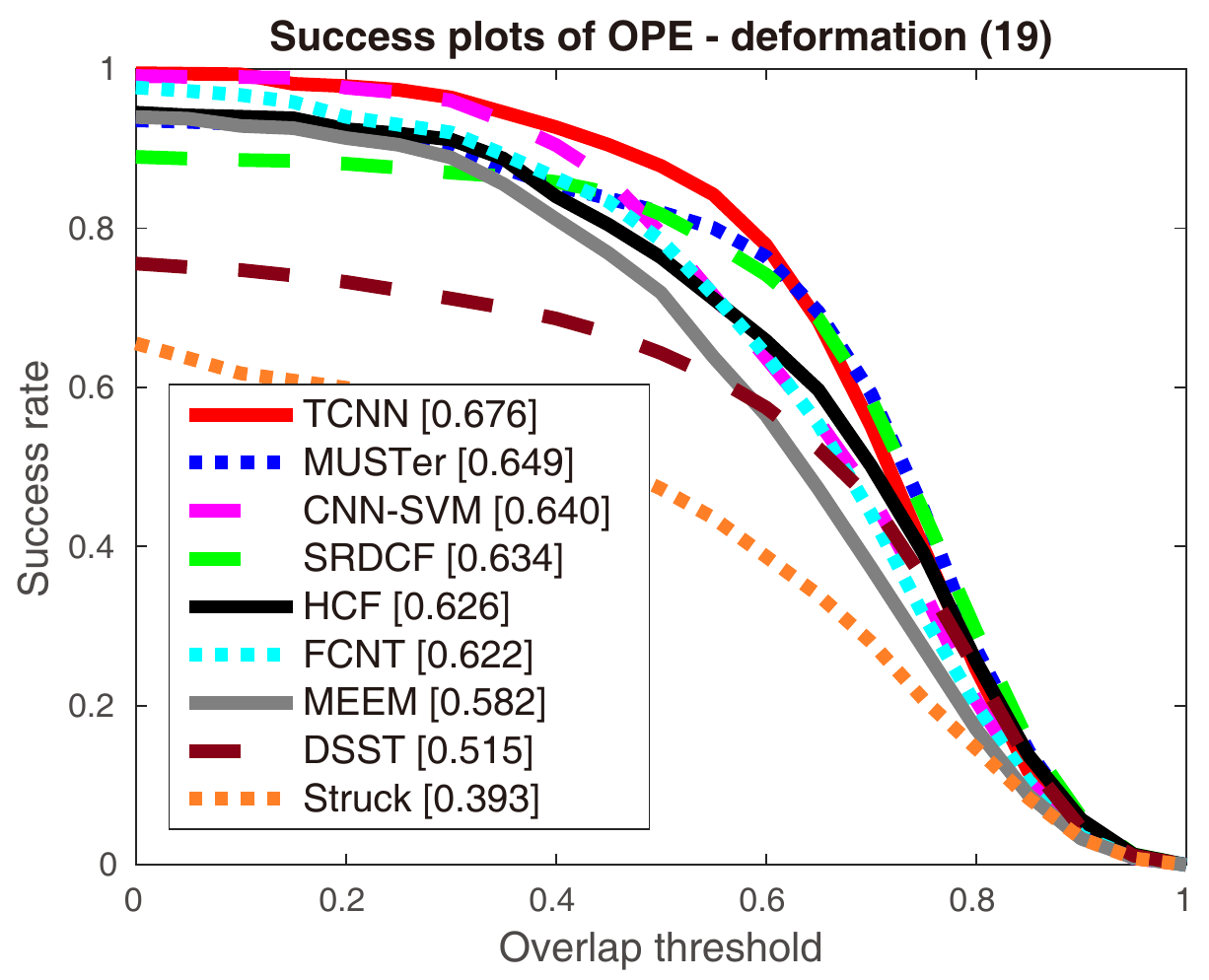}%
\includegraphics[width=0.24\linewidth]{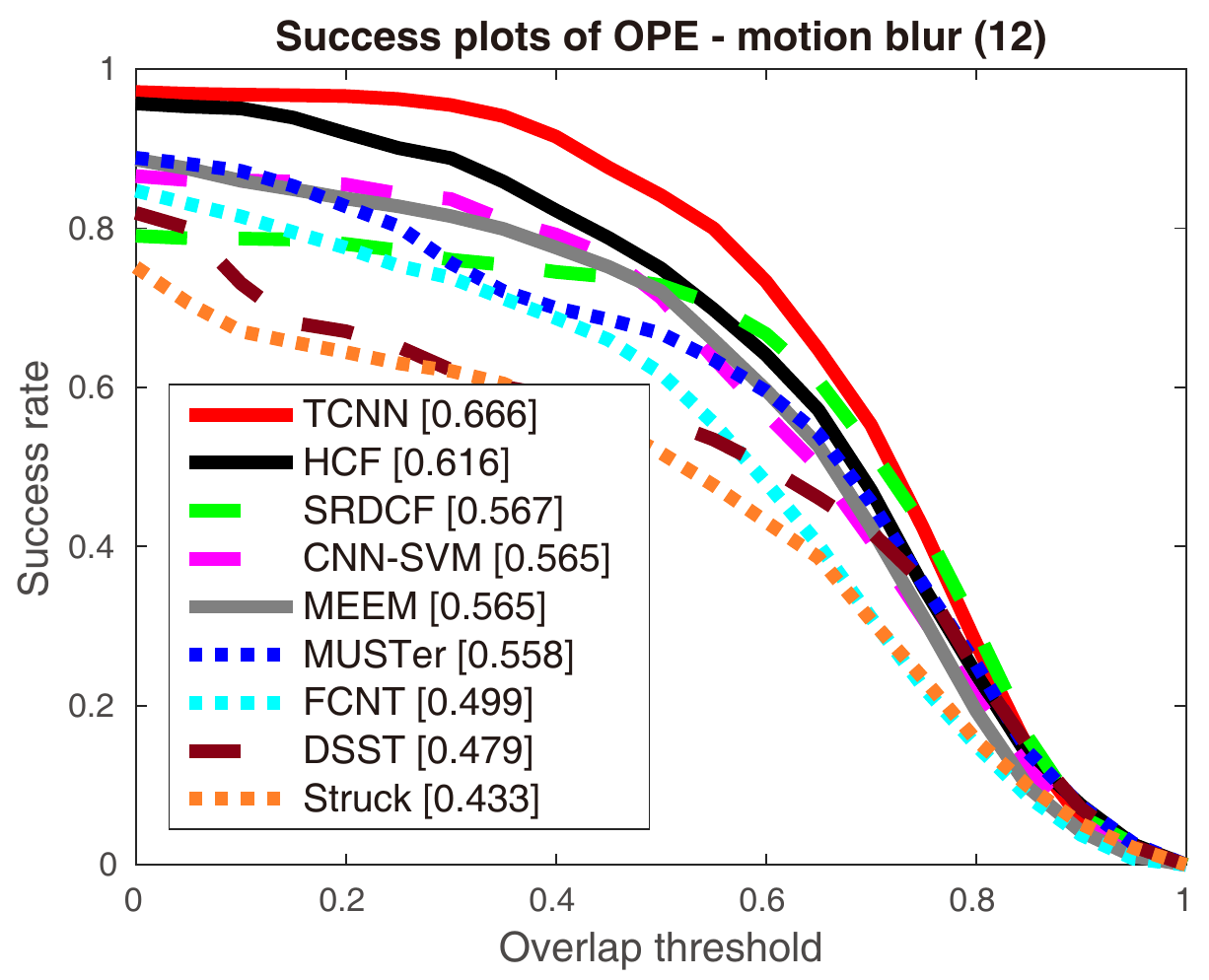}%
\includegraphics[width=0.24\linewidth]{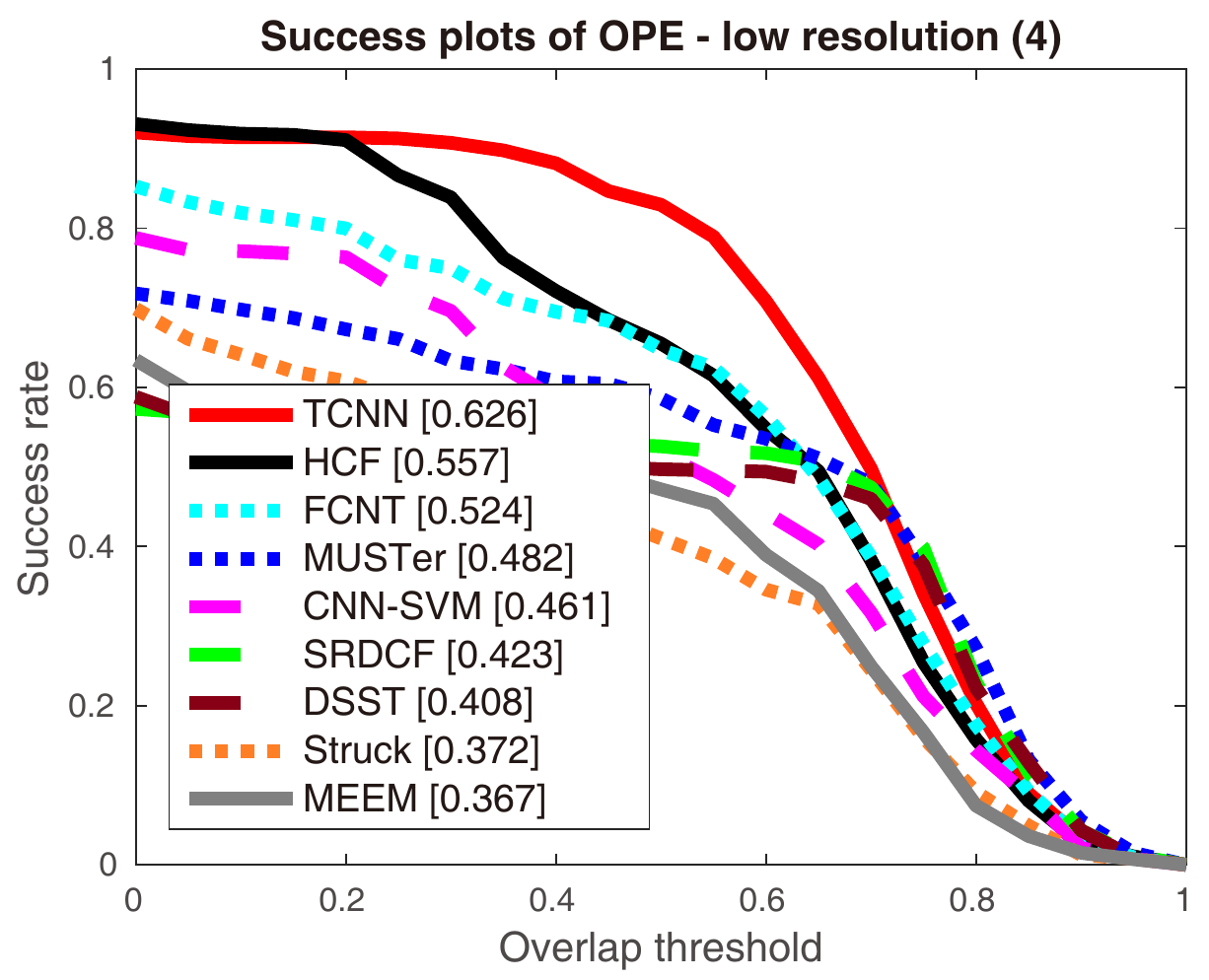} \\
\includegraphics[width=0.24\linewidth]{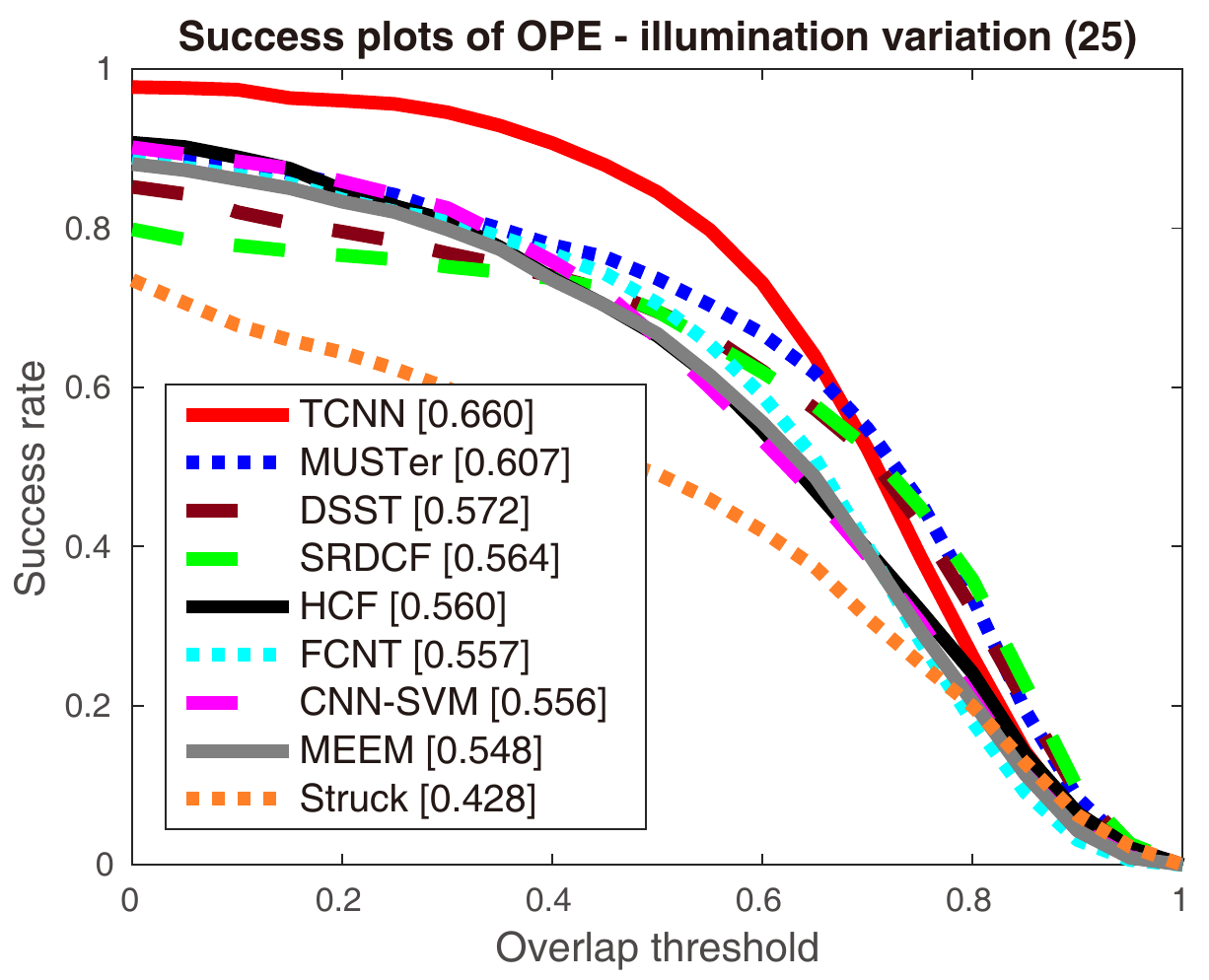}%
\includegraphics[width=0.24\linewidth]{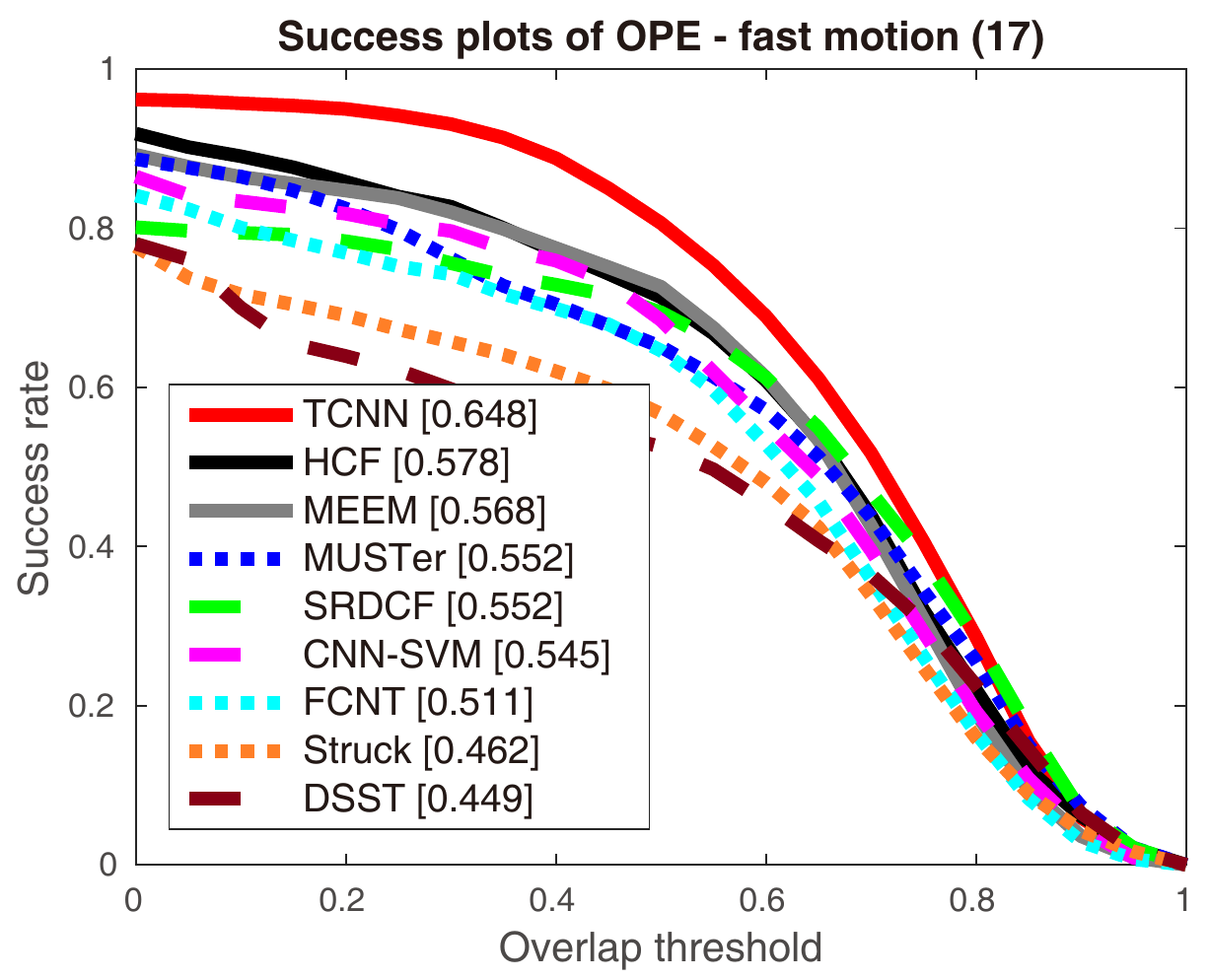}%
\includegraphics[width=0.24\linewidth]{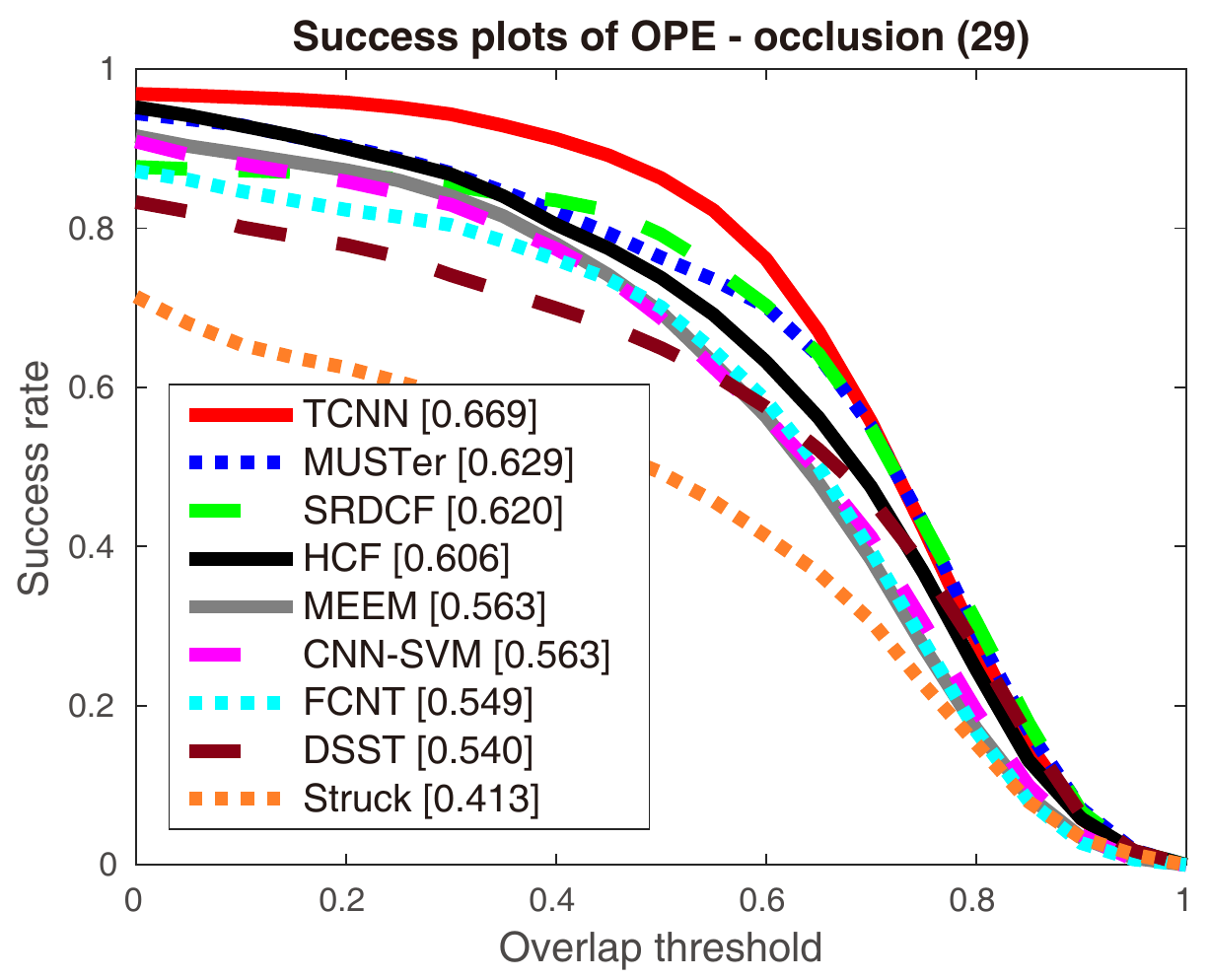}%
\includegraphics[width=0.24\linewidth]{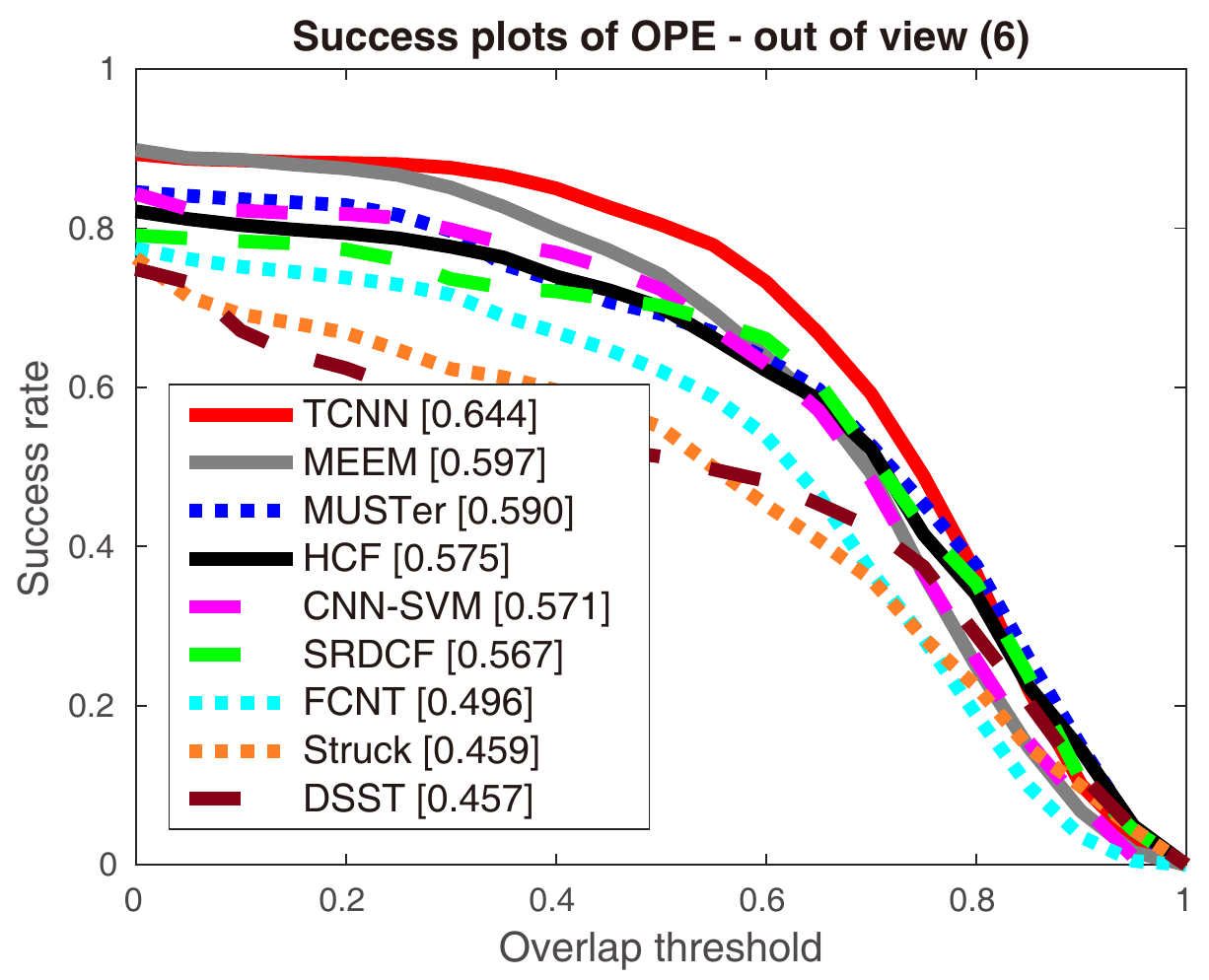}%
\end{center}
\vspace{-2mm}
\caption{Quantitative results on OTB50~\cite{otb1} for 8 challenge attributes: in-plane rotation, deformation, motion blur, low resolution, illumination variation, fast motion, occlusion, and out of view.}
\label{fig:otb_attributes}
\end{figure*}

The proposed algorithm is compared with the eight state-of-the-art trackers including  HCT~\cite{ma2015hierarchical}, CNN-SVM~\cite{hong2015online}, FCNT~\cite{wang2015visual}, SRDCF~\cite{danelljan2015learning}, MUSTer~\cite{hong2015multi}, MEEM~\cite{zhang2014meem}, DSST~\cite{danelljan2014accurate}, and Struck~\cite{hare2011struck}.
The first three algorithms employ the feature descriptors from CNNs while the rest of the methods are based on the traditional hand-crafted features.
The parameters for all trackers are fixed for all sequences during evaluation.
The precision and success plots over 50 sequences in OTB50~\cite{otb1} and 100 sequences in OTB100~\cite{otb2} are presented in Figure~\ref{fig:otb_result}.
Our tracker is denoted by TCNN, which represents the characteristics of our algorithm, observation using tree of CNNs.

The results on the two OTB datasets show that TCNN outperforms all other trackers by substantial margins.
These outstanding results are partly attributed to the strength of CNN features; the learned representations by CNNs are more effective to capture semantic information of a target than low-level hand-crafted features.
In particular, TCNN outperforms other CNN-based trackers, HCT, CNN-SVM, and FCNT, which incorporate even deeper networks like AlexNet~\cite{krizhevsky2012imagenet} or VGG-net~\cite{SimonyanICLR15}.
The better accuracy of our algorithm implies that our model update and state estimation strategy using multiple CNNs based on a tree structure is helpful to deal with various challenges. 
On the other hand, we believe that this is because the high-level features obtained from CNNs may contain insufficient spatial information and this is aggravated as the network goes deeper.
CNN-based representations are often not as effective as low-level features for the purpose of tight localization; our algorithm is slightly less accurate in the range for strict thresholds compared with MUSTer and SRDCF, which are based on correlation filters.

\begin{figure*}
\begin{center}
\includegraphics[width=0.195\linewidth]{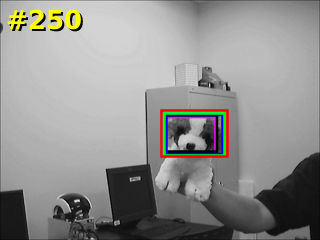}
\includegraphics[width=0.195\linewidth]{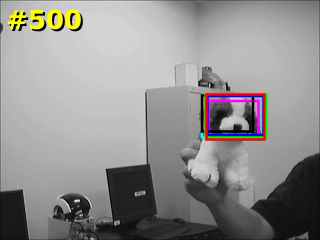}
\includegraphics[width=0.195\linewidth]{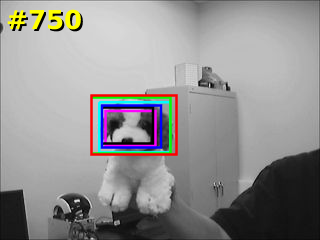}
\includegraphics[width=0.195\linewidth]{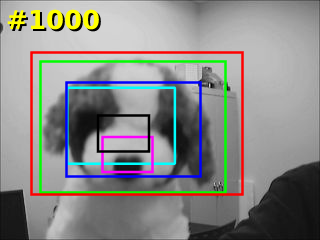}
\includegraphics[width=0.195\linewidth]{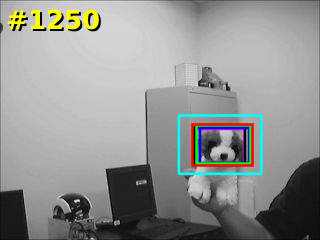}

\includegraphics[width=0.195\linewidth]{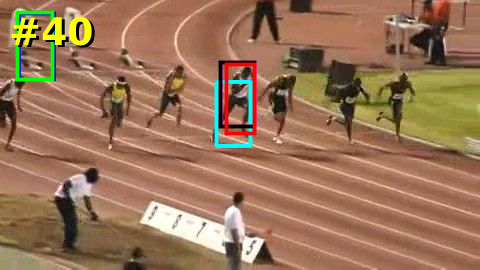}
\includegraphics[width=0.195\linewidth]{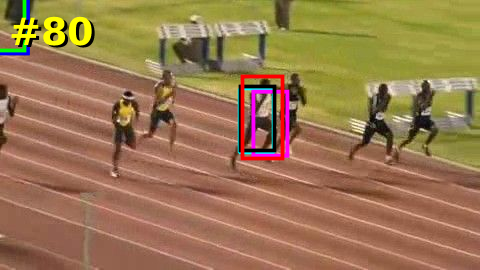}
\includegraphics[width=0.195\linewidth]{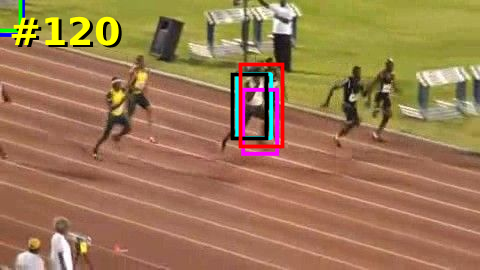}
\includegraphics[width=0.195\linewidth]{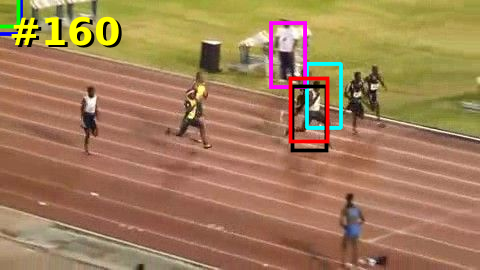}
\includegraphics[width=0.195\linewidth]{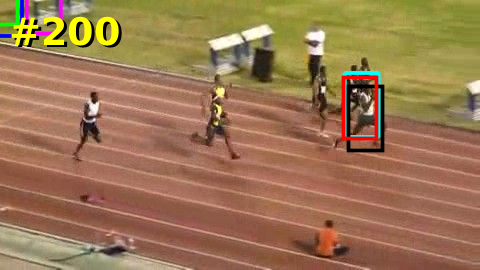}

\includegraphics[width=0.195\linewidth]{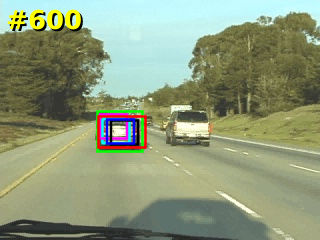}
\includegraphics[width=0.195\linewidth]{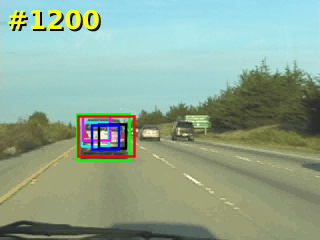}
\includegraphics[width=0.195\linewidth]{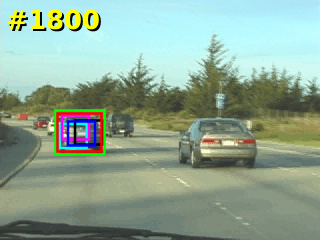}
\includegraphics[width=0.195\linewidth]{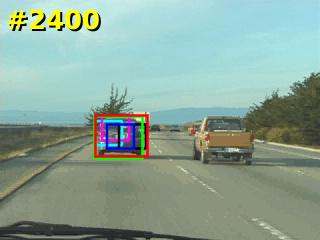}
\includegraphics[width=0.195\linewidth]{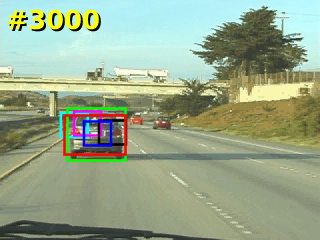}

\includegraphics[width=0.195\linewidth]{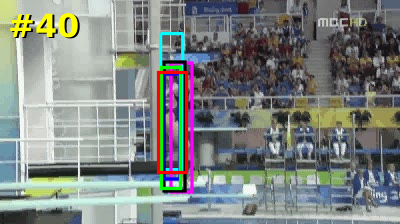}
\includegraphics[width=0.195\linewidth]{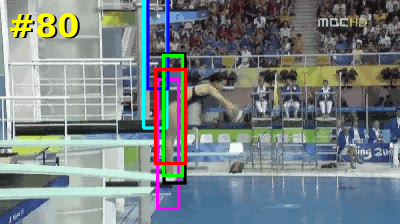}
\includegraphics[width=0.195\linewidth]{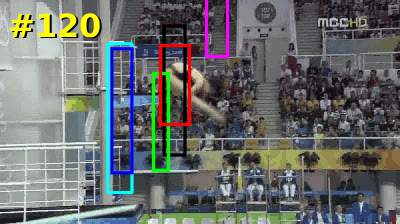}
\includegraphics[width=0.195\linewidth]{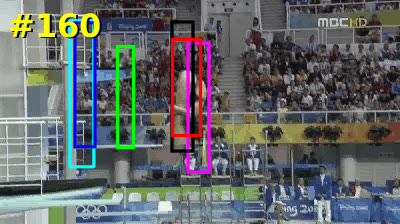}
\includegraphics[width=0.195\linewidth]{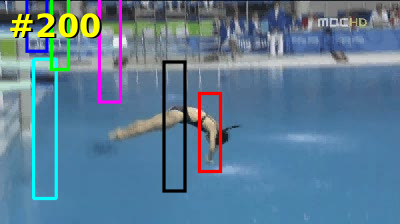}

\includegraphics[width=0.195\linewidth]{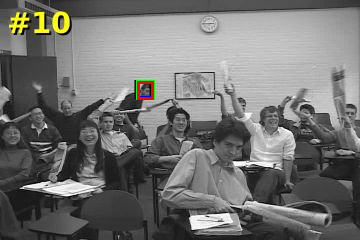}
\includegraphics[width=0.195\linewidth]{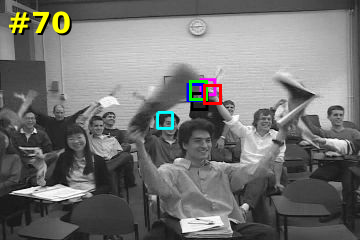}
\includegraphics[width=0.195\linewidth]{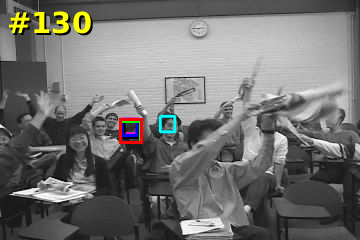}
\includegraphics[width=0.195\linewidth]{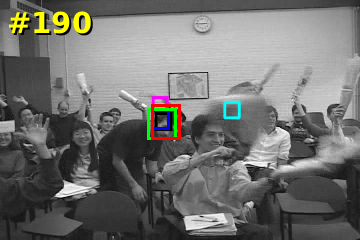}
\includegraphics[width=0.195\linewidth]{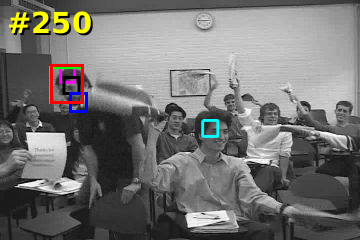}

\includegraphics[width=0.195\linewidth]{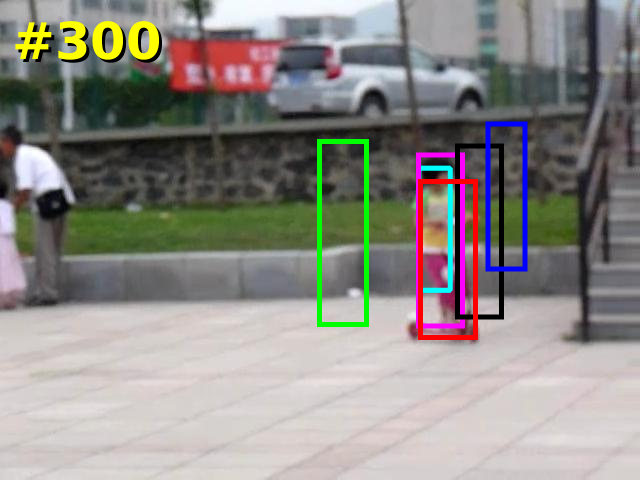}
\includegraphics[width=0.195\linewidth]{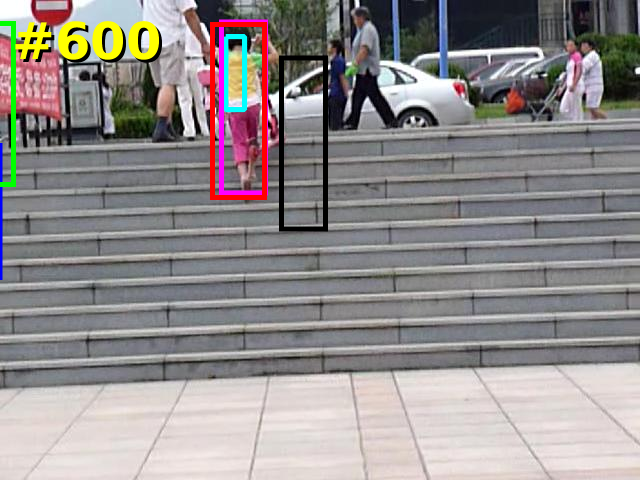}
\includegraphics[width=0.195\linewidth]{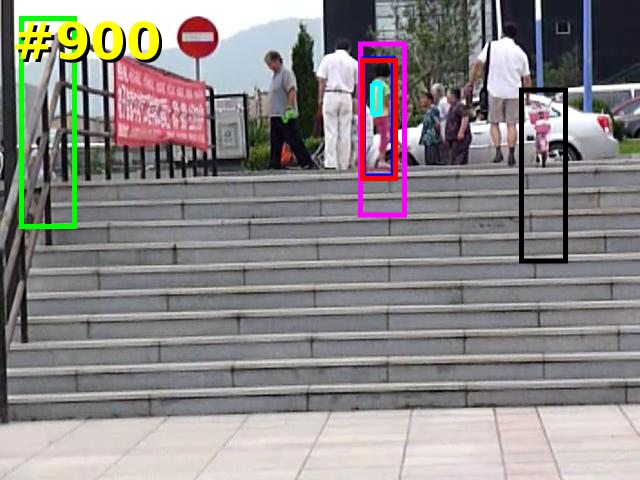}
\includegraphics[width=0.195\linewidth]{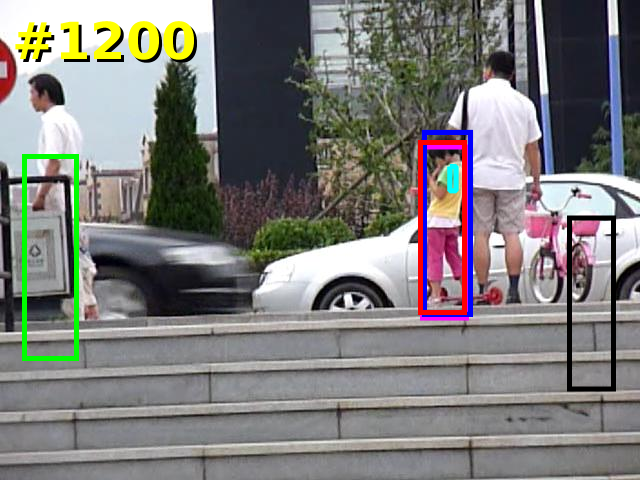}
\includegraphics[width=0.195\linewidth]{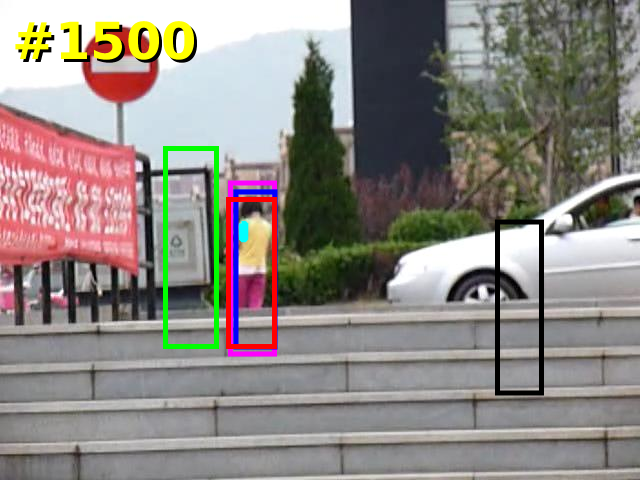}

\includegraphics[width=0.195\linewidth]{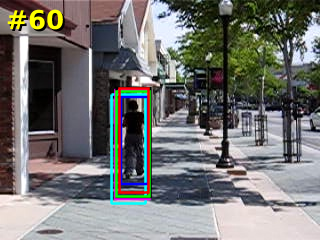}
\includegraphics[width=0.195\linewidth]{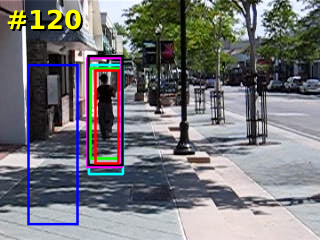}
\includegraphics[width=0.195\linewidth]{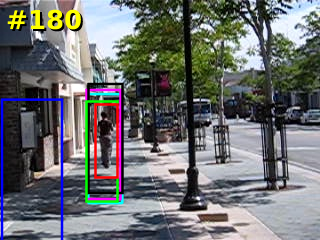}
\includegraphics[width=0.195\linewidth]{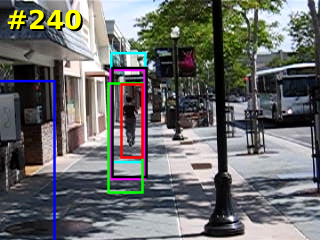}
\includegraphics[width=0.195\linewidth]{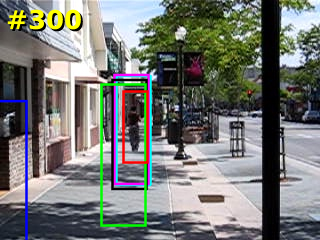}

\includegraphics[width=0.195\linewidth]{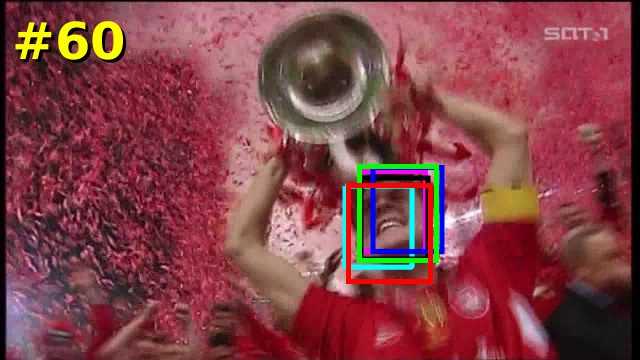}
\includegraphics[width=0.195\linewidth]{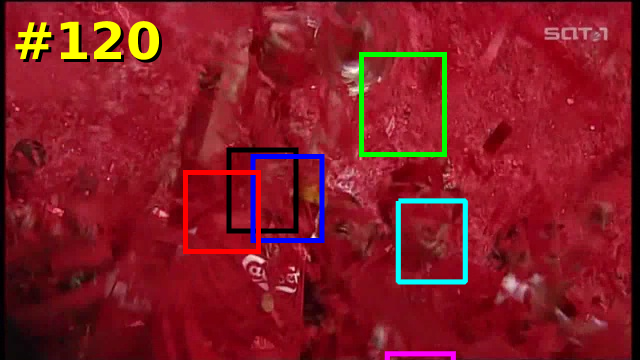}
\includegraphics[width=0.195\linewidth]{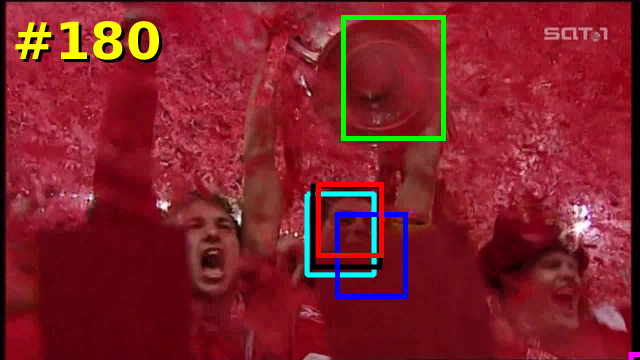}
\includegraphics[width=0.195\linewidth]{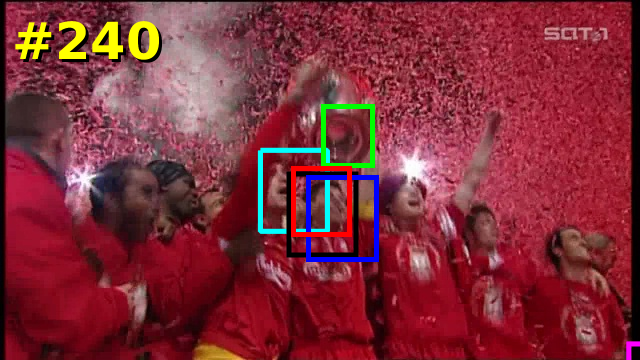}
\includegraphics[width=0.195\linewidth]{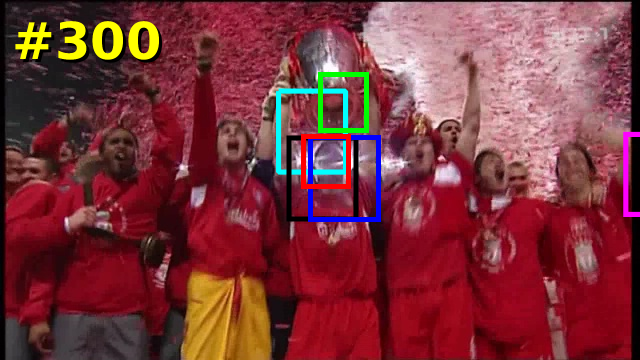}

\includegraphics[width=0.195\linewidth]{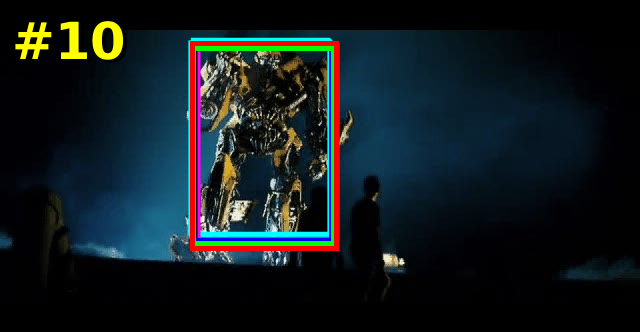}
\includegraphics[width=0.195\linewidth]{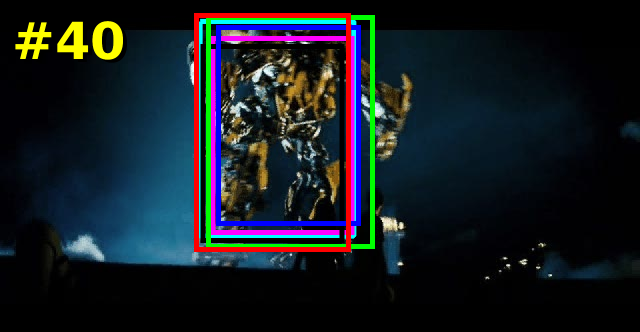}
\includegraphics[width=0.195\linewidth]{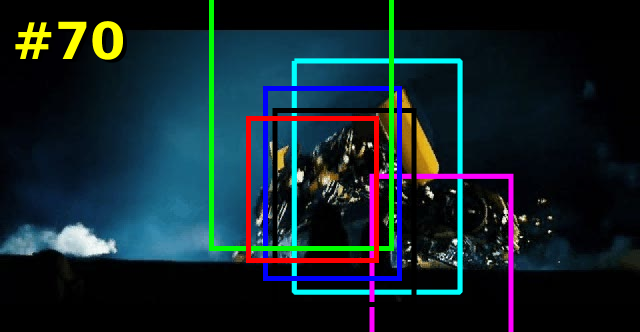}
\includegraphics[width=0.195\linewidth]{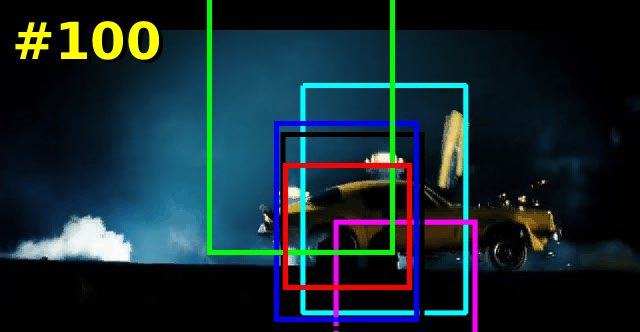}
\includegraphics[width=0.195\linewidth]{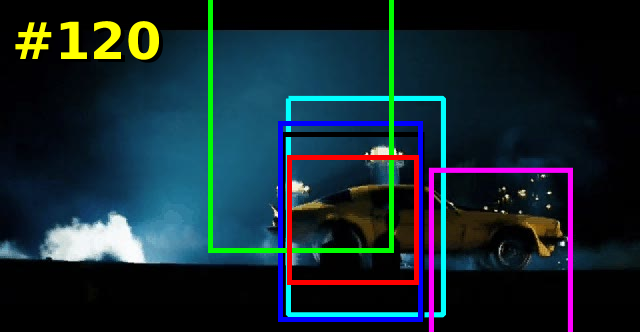}

\includegraphics[width=0.95\linewidth]{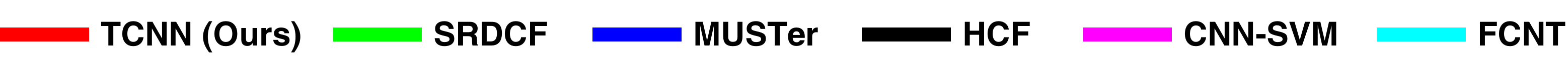}

\end{center}
\vspace{-5mm}
\caption{Qualitative results in several challenging sequences of OTB100 dataset (\emph{Dog1}, \emph{Bolt2}, \emph{Car24}, \emph{Diving}, \emph{Freeman4}, \emph{Girl2}, \emph{Human9}, \emph{Soccer}, and \emph{Trans}).}
\label{fig:qualOTB}
\end{figure*}

Figure~\ref{fig:otb_attributes} presents the performance of tracking algorithms for various challenging attributes provided in the benchmark; TCNN is effective in handling all kinds of attributes compared to the existing state-of-the-art methods.
The trackers based on low-level features generally fail to track targets in challenging situations such as rotation, deformation, motion blur and low resolution, which require high-level understanding about targets.
The approaches based on CNNs work well in general, but are not successful typically in illumination variations and occlusion probably due to dramatic changes in target appearances.
TCNN is not outstanding in dealing with out-of-view situation, because it is based on local candidate sampling and does not incorporate any re-detection module.
The qualitative tracking results by multiple algorithms on a subset of sequences are illustrated in Figure~\ref{fig:qualOTB}.

\subsubsection{Internal Analysis}
\begin{table}[t]
\caption{Internal comparison results. The rates of successfully tracked frames in terms of center location error with threshold 20 pixels and bounding box overlap ratio with threshold 0.5 are denoted by precision and success, respectively. AUC is based on the success plot in OTB evaluation protocol~\cite{otb1,otb2}.}
\label{tab:internal}
\small
\begin{center}
\begin{tabular}{c||c|c|c}
& ~Precision (\%)~& ~Success (\%)~& ~AUC (\%)~ \\ \hline
~{Linear\_single}~ & 89.6 & 85.8 & 65.8 \\
~{Linear\_mean}~ & 92.0 & 86.9 & 67.2 \\
~{Tree\_mean}~ & 92.8 & 86.8 & 67.4 \\
~{Tree\_max}~  & 92.0 & 87.0 & 67.2 \\ \hline
~TCNN~ & \textbf{93.7} & \textbf{87.9} & \textbf{68.2} \\ \hline
\end{tabular}
\end{center}
\vspace{-5mm}
\end{table}

\begin{figure*}[t]
\begin{center}
\includegraphics[width=0.85\linewidth]{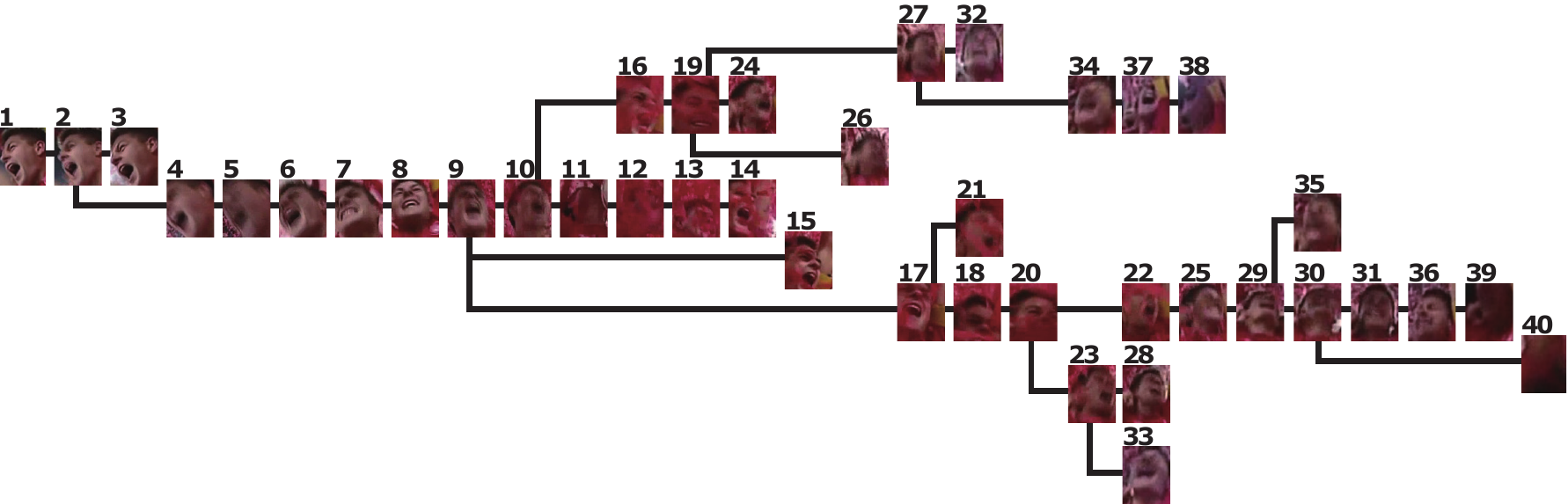}
\end{center}
\vspace{-5mm}
\caption{The example of identified tree structure by our algorithm in \textit{Soccer} sequence. Each vertex $v$ is associated with a CNN, and the estimated target in the representative frames in $\mathcal{F}_v$ is shown for each vertex.  The number for each vertex denotes the index of the CNN in the tree structure.  Our algorithm maintains multiple paths, through which target appearance changes smoothly to improve the reliability of each model and the diversity of overall models.}
\label{fig:graph}
\end{figure*}

To analyze the effectiveness of our algorithm, we evaluate the variations of our tracker.
We implemented the following four additional versions: a single CNN with sequential updates ({Linear\_single}), multiple CNNs with state estimation by simple averaging and sequential updates ({Linear\_mean}), multiple CNNs with state estimation by simple averaging and tree updates ({Tree\_mean}), and multiple CNNs with state estimation by maximum weight and tree updates  ({Tree\_max}).
Table~\ref{tab:internal} summarizes the results from the internal comparison as follows.
First, using multiple models (Linear\_mean) is more effective than using a single model (Linear\_single) because of the benefit from model diversity. 
Second, the approaches maintaining CNNs in a tree structure (Tree\_mean) enhances performance compared to the method based on a single path (Linear\_mean) since it allows each CNN to be updated more reliably through more appropriate paths. 
Finally, the proposed algorithm (TCNN) is more effective than all the other variations, which verifies that both our model update and state estimation strategies are useful to improve performance.

We tested the impact of bounding box regression on the quality of our tracking algorithm.
The representative accuracy in precision and success of TCNN without bounding box regression are (0.923, 0.669) in OTB50, and (0.870, 0.634) in OTB100.
According to these results, bounding box regression is a useful component to alleviate the limitation of CNN-based features in object localization.

%visualize
To visualize the advantage of our tree-based model maintenance technique, we present an example of identified tree structure in Figure~\ref{fig:graph}, where vertices correspond to estimated target bounding boxes.
It demonstrates that the tree structure maintains multiple reliable paths while isolating the frames with significantly different target appearances due to heavy occlusion or other challenges in a local branch to avoid drift problems.

\subsection{Evaluation on VOT2015 Dataset}

We also tested the proposed algorithm in the recent dataset for visual object tracking 2015 challenge (VOT2015)~\cite{vot15}, which is composed of 60 challenging video sequences with sufficient variations. 
The VOT challenge provides re-initialization protocol, where trackers are reset with ground-truths in the middle of evaluation if tracking failures are observed.
The performance metrics are defined based on accuracy and robustness, which are computed by the bounding box overlap ratio and the number of tracking failures.
The VOT challenge also introduces the expected average overlap as a new metric to rank tracking algorithms; it estimates how accurate the estimated bounding box is after a certain number of frames are processed since initialization.

TCNN is compared with 12 trackers, which do not include three CNN-based algorithms used for the evaluation in OTB since their results are not available in VOT2015 dataset while a few good performing algorithms in the dataset, \ie EBT~\cite{gao2015tracking} and DeepSRDCF~\cite{danelljan2015convolutional}, are included for comparison.\footnote{A few algorithms that have additional training procedure are not included in our experiment.}
In addition, we include all algorithms that are tested on OTB but do not rely on CNNs, \eg SRDCF, DSST, MUSTer, MEEM and Struck, in this evaluation.

\begin{table}[t]
\small
\centering
\caption{The average scores and ranks of accuracy and robustness on the experiment in VOT2015~\cite{vot15}. The first and second       
best scores are highlighted in red and blue colors, respectively.}
\label{table:vot}
\resizebox{\linewidth}{!}{
\begin{tabular}{c||c c|c c|c}
\multirow{2}{*}{{Trackers}}					& \multicolumn{2}{c|}{{Accuracy}}			& \multicolumn{2}{c|}{{Robustness}}			& {Expected}			\\ \cline{2-5}
										& {Rank}				& {Score}				& {Rank}				& {Score}				& {overlap ratio}		\\ \hline
{DSST~\cite{danelljan2014accurate}}			& 3.48				& 0.54				& 7.93				& 2.56				& 0.1719				\\
{MUSTer~\cite{hong2015multi}}				& 3.42				& 0.52				& 6.13				& 2.00				& 0.1948				\\
{MEEM~\cite{zhang2014meem}}					& 4.08				& 0.50				& 6.02				& 1.85				& 0.2212				\\
{Struck~\cite{hare2011struck}}				& 5.27				& 0.47				& 4.05				& 1.64				& 0.2389				\\
{RAJSSC~\cite{zhang2016joint}}				& 2.08				& {\color{blue}0.57}	& 4.87				& 1.63				& 0.2420				\\
{NSAMF~\cite{Li2015scale}}					& 3.22				& 0.53				& 3.85				& 1.29				& 0.2536				\\
{SC-EBT~\cite{wang2014ensemble}}			& 2.27				& 0.55				& 5.07				& 1.90				& 0.2540				\\
{sPST~\cite{hua2015online}}					& 2.78				& 0.55				& 4.67				& 1.48				& 0.2767				\\
{LDP~\cite{LukezicCK16}}					& 4.58				& 0.49				& 4.65				& 1.33				& 0.2785				\\
{SRDCF~\cite{danelljan2015learning}}			& 2.32				& 0.56				& 3.48				& 1.24				& 0.2877				\\
{EBT~\cite{gao2015tracking}}				& 6.30				& 0.48				& {\color{red}2.75}	& 1.02				& 0.3160				\\
{DeepSRDCF~\cite{danelljan2015convolutional}}	& 2.23				& 0.56				& 2.90				& 1.05				& 0.3181				\\ \hline
{TCNN w/o BBR}							& {\color{blue}1.83}	& 0.57				& 3.30				& {\color{blue}0.75}	& {\color{blue}0.3321}	\\ 
{TCNN}									& {\color{red}1.58}	& {\color{red}0.59}	& {\color{blue}2.83}	& {\color{red}0.74}	& {\color{red}0.3404}	\\ \hline
\end{tabular}
}
\end{table}

Table~\ref{table:vot} illustrates the strength of TCNN in VOT2015 dataset compared to other trackers.
TCNN outperforms all the compared algorithms in most of evaluation metrics while DeepSRDCF~\cite{danelljan2015convolutional} and EBT~\cite{gao2015tracking} shows comparable results.
DeepSRDCF is the improved version of SRDCF~\cite{danelljan2015learning} with a deep feature, but the performance gain is not substantial.
EBT is robust since the tracker finds target in the entire image rather than a local search window, however, its overall score in terms of expected overlap ratio is relatively low since it is not good at localization.
Comparing these algorithms, TCNN achives better performance by fully utilizing the representation power of deep features from multiple models.
TCNN also outperforms DeepSRDCF and EBT in OTB50; according to \cite{danelljan2015convolutional} and \cite{gao2015tracking}, the success rate of DeepSRDCF and EBT is 0.649 and 0.581, respectively, while that of TCNN is 0.682.
Note that bounding box regression turns out to be helpful in VOT2015 dataset as well.
It is interesting that MUSTer is not as competitive as in the OTB~\cite{otb1,otb2} while TCNN is outstanding in both OTB and VOT2015 datasets consistently compared to other trackers.

%---------------------------------------------------------------------------------------------------------------------------------------------------------
% CONCLUSION
%---------------------------------------------------------------------------------------------------------------------------------------------------------
\section{Conclusion}
\label{sec:conclusion}
We presented a novel tracking algorithm based on multiple CNNs maintained in a tree structure, which are helpful to achieve multi-modality and reliability of target appearances.
The state estimation of target is performed by computing a weighted average of scores from the multiple CNNs.
The contribution of each CNN for target state estimation and online model update is also determined by exploiting the tree structure.
Due to parameter sharing in convolutional layers, the use of multiple CNNs does not require substantial increase of memory and computation.
Our tracking algorithm outperforms the state-of-the-art techniques in both OTB and VOT2015 benchmarks.

{\small
\bibliographystyle{ieee}
\bibliography{arxiv_tcnn}
}

\end{document}